\documentclass[fleqn,10pt]{wlscirep}
\usepackage[utf8]{inputenc}
\usepackage[T1]{fontenc}
\usepackage{wrapfig}
\usepackage{lineno}

\title{Automatic Cell Counting in Flourescent Microscopy Using Deep Learning}

\author[1,2,*]{Roberto Morelli}
\author[1,2]{Luca Clissa}
\author[4]{Marco Dalla}
\author[3]{Marco Luppi}
\author[1,2]{Lorenzo Rinaldi}
\author[1,2]{Antonio Zoccoli}
\affil[1]{Istituto Nazionale di Fisica Nucleare}
\affil[2]{University of Bologna, Department of Physics and Astronomy, Bologna, Italy}
\affil[3]{University of Bologna, Department of Biomedical and Neuromotor Sciences, Bologna, Italy}
\affil[4]{University College, Department of Computer Science, Cork, Ireland}


\begin{abstract}
Counting cells in fluorescent microscopy is a tedious, time-consuming task that researchers have to accomplish to assess the effects of different experimental conditions on biological structures of interest. Although such objects are generally easy to identify, the process of manually annotating cells is sometimes subject to arbitrariness due to the operator's interpretation of the borderline cases.

We propose a Machine Learning approach that exploits a fully-convolutional network in a binary segmentation fashion to localize the objects of interest. Counts are then retrieved as the number of detected items.

Specifically, we adopt a UNet-like architecture leveraging residual units and an extended bottleneck for enlarging the field-of-view. In addition, we make use of weighted maps that penalize the errors on cells boundaries increasingly with overcrowding.
These changes provide more context and force the model to focus on relevant features during pixel-wise classification. As a result, the model performance is enhanced, especially in presence of clumping cells, artifacts and confounding biological structures.
Posterior assessment of the results with domain experts confirms that the model detects cells of interest correctly. The model demonstrates a human-level ability inasmuch even erroneous predictions seem to fall within the limits of operator interpretation.
This qualitative assessment is also corroborated by quantitative metrics as an ${F_1}$ score of 0.87.

Despite some difficulties in interpretation, results are also satisfactory with respect to the counting task, as testified by mean and median absolute error of, respectively, 0.8 and 1.

\end{abstract}

\begin{document}

\flushbottom
\maketitle
%
%
\thispagestyle{empty}



\section{Introduction}
\label{intro}
There is no doubt Deep Convolutional Neural Networks have shown the ability to outperform the state-of-the-art in many computer vision applications in the past decade. Successful applications range from classification and detection of basically any kind of objects \cite{AlexNet, YOLO} to segmentation of tumors in medical imaging \cite{brain_tumor, breast_cancer, ciresan2012deep, cirecsan2013mitosis} and generative models for image reconstruction \cite{reconstruction} and super-resolution \cite{super-resolution}.

This work tackles the problem of counting cells into biological images, also providing a justification of the output number by means of a segmentation map that localizes the detected objects. 
This additional information is particularly relevant to support the results with a clear, visual evidence of which objects contribute to the final counts.

Counting objects in digital images is a common task for many real-world applications which is sometimes still performed manually by human annotators. However, this process may be very time-consuming depending on the number of available images. Also, the task becomes tedious when the objects to count appear in large quantities, thus leading to errors due to fatigue of the operators.
On the other hand, sometimes there are images in which either the true objects are not clearly identifiable or background looks very similar to the structures of interest. When that is the case, the results become arguable and subjective due to the interpretation of such borderline cases.

Therefore, the introduction of automatic procedures to detect and count objects in digital images would bring many benefits in such applications, speeding up the operations and lightening the efforts of researchers, as well as limiting fatigue errors and standardizing to the systematic effect of the model the arbitrariness due to multiple operators influence.

\subsection{Background and motivation}
\label{sec:back_mot}
Recently, Hitrec et al. \cite{hitrec2019neural} investigated the brain areas of mice that mediate the entrance into torpor and showed evidence of which networks of neurons are associated with this process.
Knowing and controlling the mechanisms that rule the onset of lethargy may have a significant impact when coming to applications to humans.
Being able to artificially induce hibernation may be crucial for a wide variety of medical applications, from intensive care to oncology, as well as space travels and more.
As a consequence,  their work arouses considerable interest on the topic and lays the foundations for novel and more in-depth studies.

However, such an experiment requires an operator to inspect hundreds of pictures of brain slices manually and annotate the cells of interest (stained cells) in each image. The drawback is that the procedure is time-consuming and prone to errors due to fatigue. Also, despite target structures are generally clearly identifiable and little practice is needed to train a new operator for the task, there is still a certain degree of intrinsic arbitrariness due to the operator's interpretation of borderline cases.

For these reasons, our work aims at facilitating and speeding up future research in this and similar fields through the adoption of an intelligent system that counts the objects of interest without human intervention.

The advantages of doing so are two-fold. 
On one side, the benefit in terms of time and human effort saved through the automation of the task is evident.
On the other, using a Machine Learning (ML) model for different experiments would introduce a systematic ``operator" effect, thus limiting the arbitrariness of borderline cases both within and between experiments.

\subsection{Task}
\label{sec:task}
Different approaches to count objects in an image have been proposed in the literature over the years. 
Indeed, this task can be carried out under multiple paradigms depending on the specific needs of the study and the available data.

The natural setting to tackle this problem is the so-called \textit{counting-by-regression} scheme. In this case the input data consist of the image, possibly allowing for other features. The model is then trained to compute the raw count of objects directly. However, this does not provide any immediate justification of which elements generated the final count.

Another approach is \textit{counting-by-detection}, in which the model is trained to reproduce ground-truth masks having bounding boxes surrounding the objects to detect. In this way, the output becomes an image in which pixels are classified into either signal (within the boxes) or background (outside) class. 
This outcome provides the raw count of objects as the number of sets of connected pixels, and a justification in terms of the localization furnished by the bounding boxes. 

Building on the latter framework, one can refine the model ability to detect and localize the objects by including pixel-wise semantic labels in the ground-truth masks. This produces a classification at a pixel level that makes it possible to distinguish also the exact boundaries of each object. The total number of objects is then retrieved again by looking at groups of connected pixels. Such an approach is referred to as \textit{counting-by-segmentation}.

Given the ultimate goal of reproducing counts of cells in each picture, the natural approach would be to consider a regression model. However, this would give poor justification (or not at all) of which are the items that contribute to the count.

For this reason -- and considering that, to the best of our knowledge, there are no attempts of using ML models in this specific domain of application --  we opted for a \textbf{counting-by-segmentation} approach. Such a framework, in fact, leverages ground-truth maps as labels during training in order to produce also a justification of the counts in terms of localization of the detected objects.

\subsection{Related works}
\label{sec:rel_works}
Lempitsky et al. \cite{lempitsky2010learning} were among the first to investigate automated approaches to count cells in digital images, and they showed how to get the total number of objects in an image starting from dot annotations. 
To this end, they introduced the concept of a density map as target. In particular, they adopted a simple linear regression model to reproduce a mixture of bidimensional Gaussians centered on dot annotations. The count was then retrieved by integrating over the predicted density map delivered by the model.

Another remarkable result was obtained by Seguí et al. \cite{segui2015learning}, who trained a convolutional neural network in a regression framework. Interestingly, the authors were able to show that the features learned during this process were also useful for object detection, meaning that the model actually learned what it was counting.

Arteta et al. \cite{arteta2016counting} successively discussed new approaches past the density map that focus on incorporating multiple dot annotations from noisy crowdsourced data, i.e. data annotated voluntarily by non-experts.

More recently, Cohen et al. \cite{paul2017count} presented a work in which they adopted a fully convolutional network derived from the Inception family 
\cite{szegedy2016rethinking}. In particular, they extended the idea of counting everything inside the receptive field as in \cite{segui2015learning}  by introducing redundant counts to make results more stable and reliable, thus transforming the concept of density map in \cite{lempitsky2010learning} into that of a count map.
\section{Dataset}
\label{sec:dataset}
\begin{figure}
\centerline{
\includegraphics[width=0.3\textwidth]{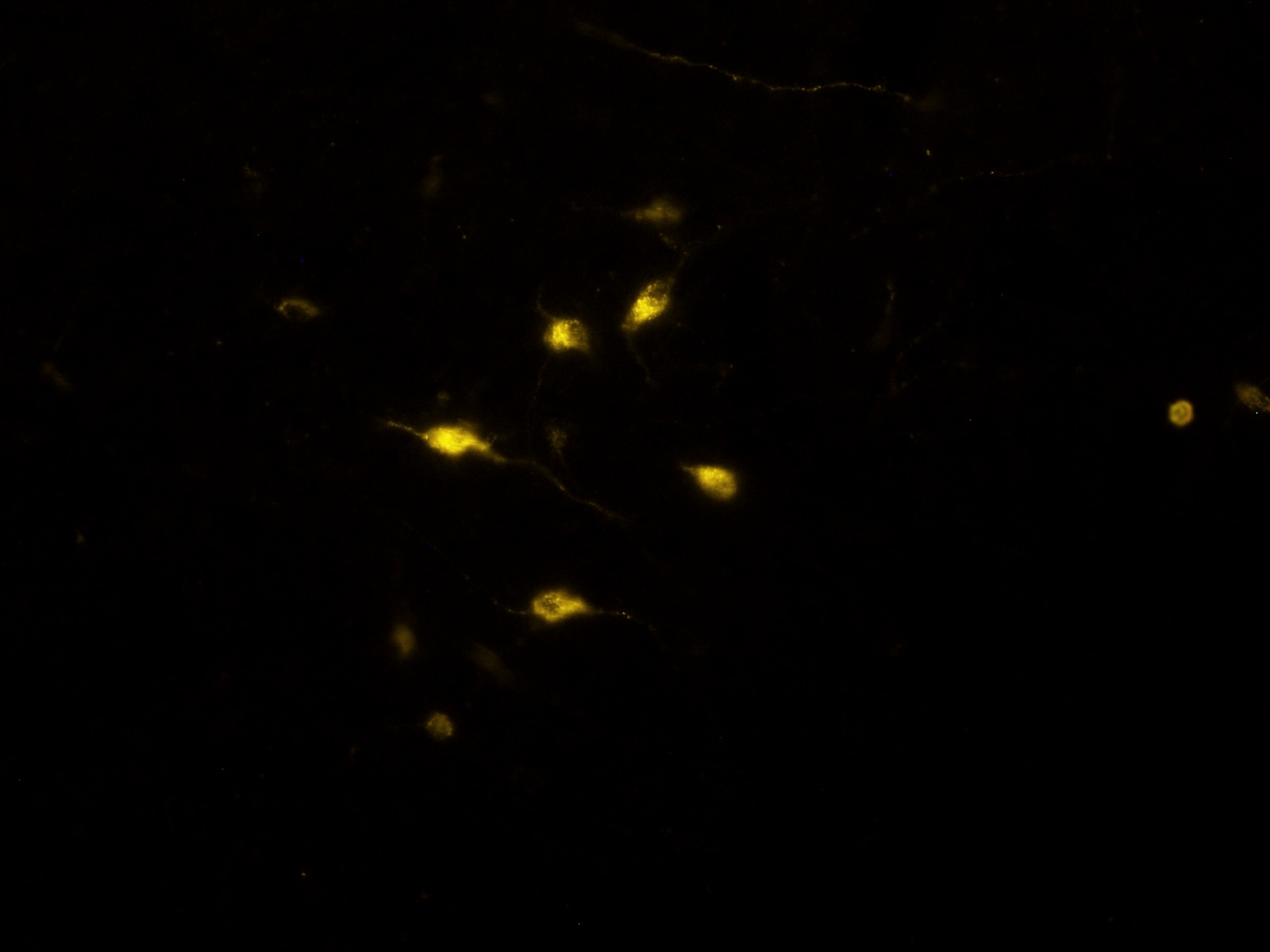}
\includegraphics[width=0.3\textwidth]{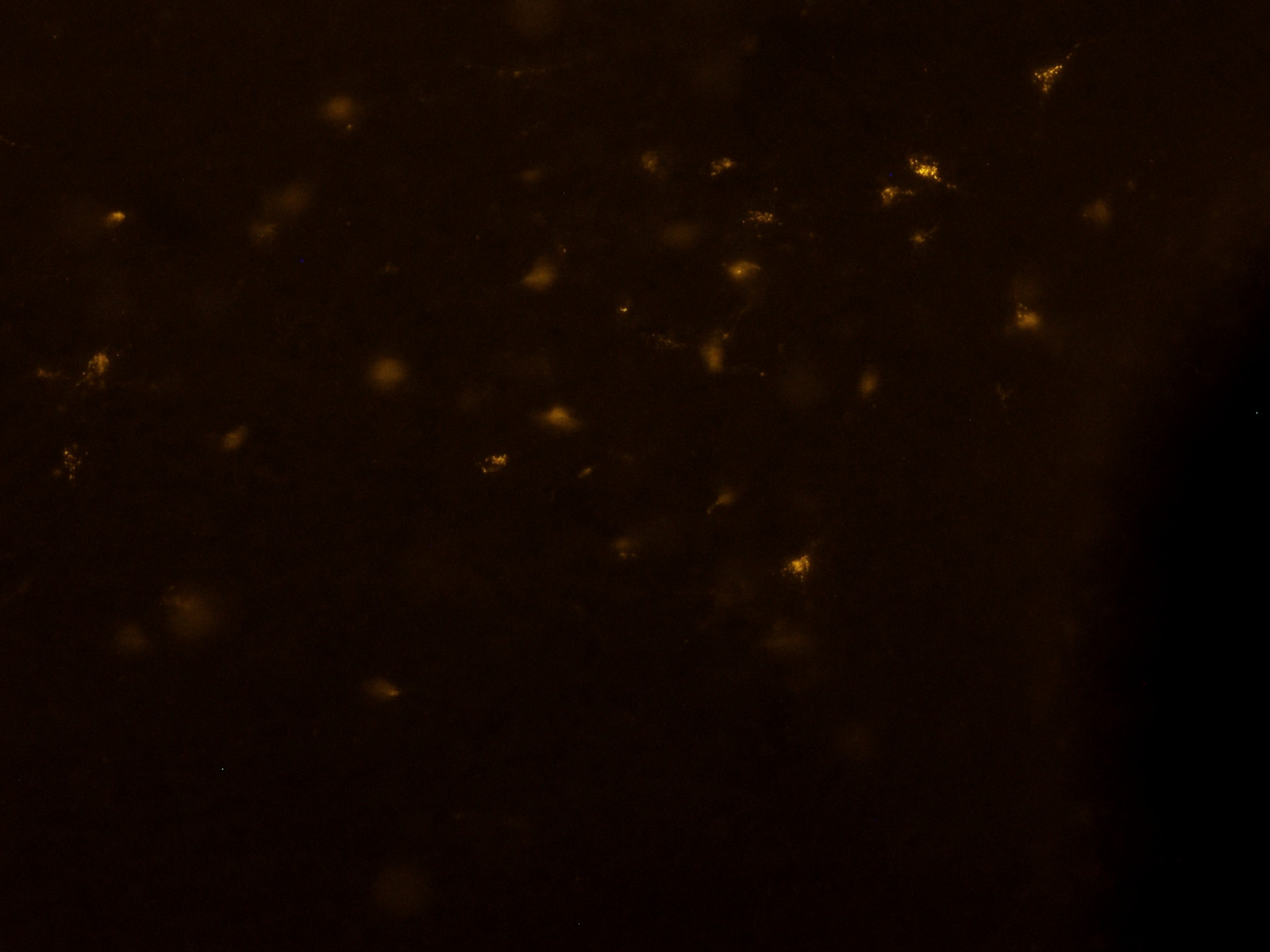}
\includegraphics[width=0.3\textwidth]{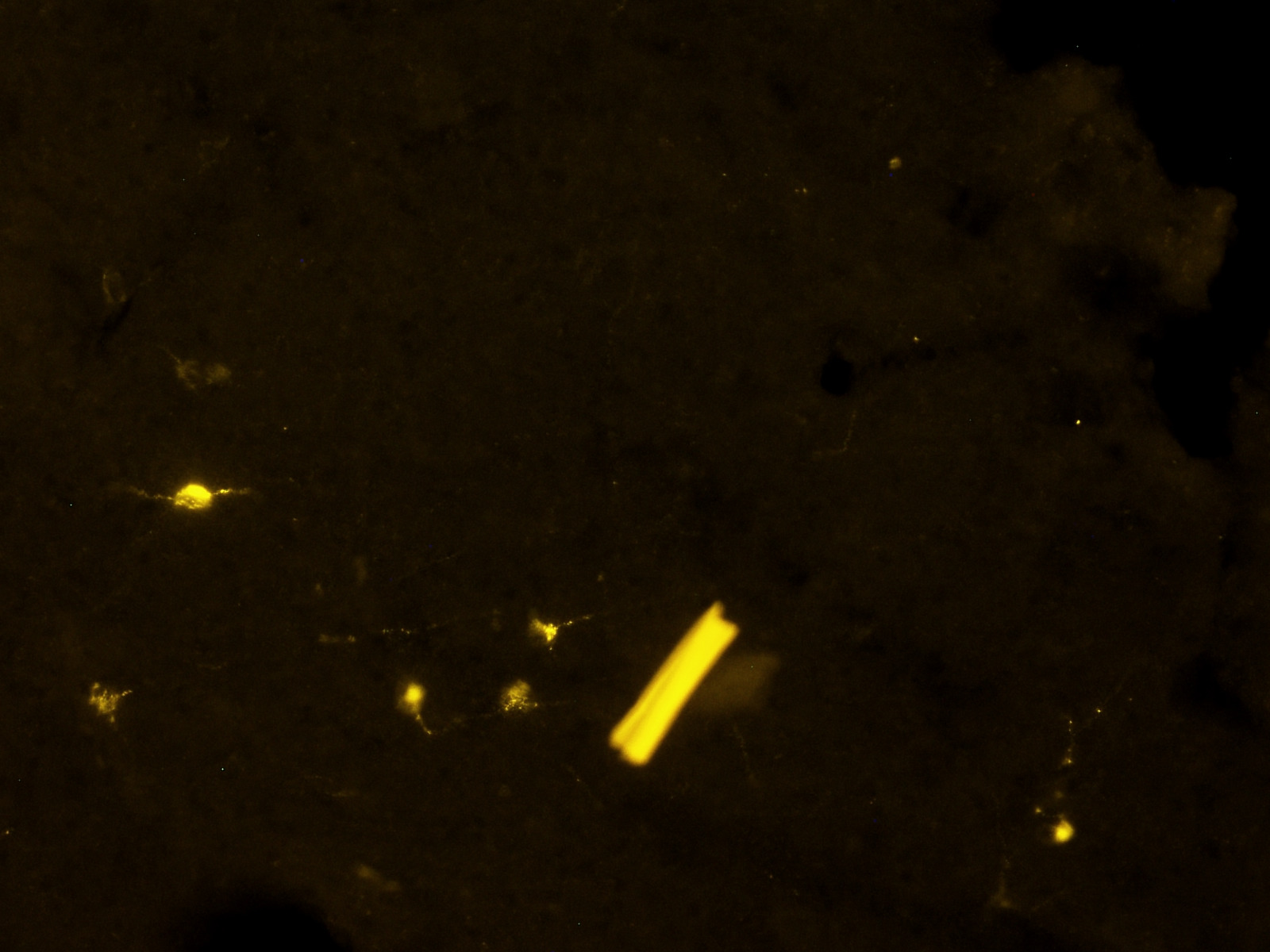}
}
\vspace{0.1cm}
\centerline{
\includegraphics[width=0.3\textwidth]{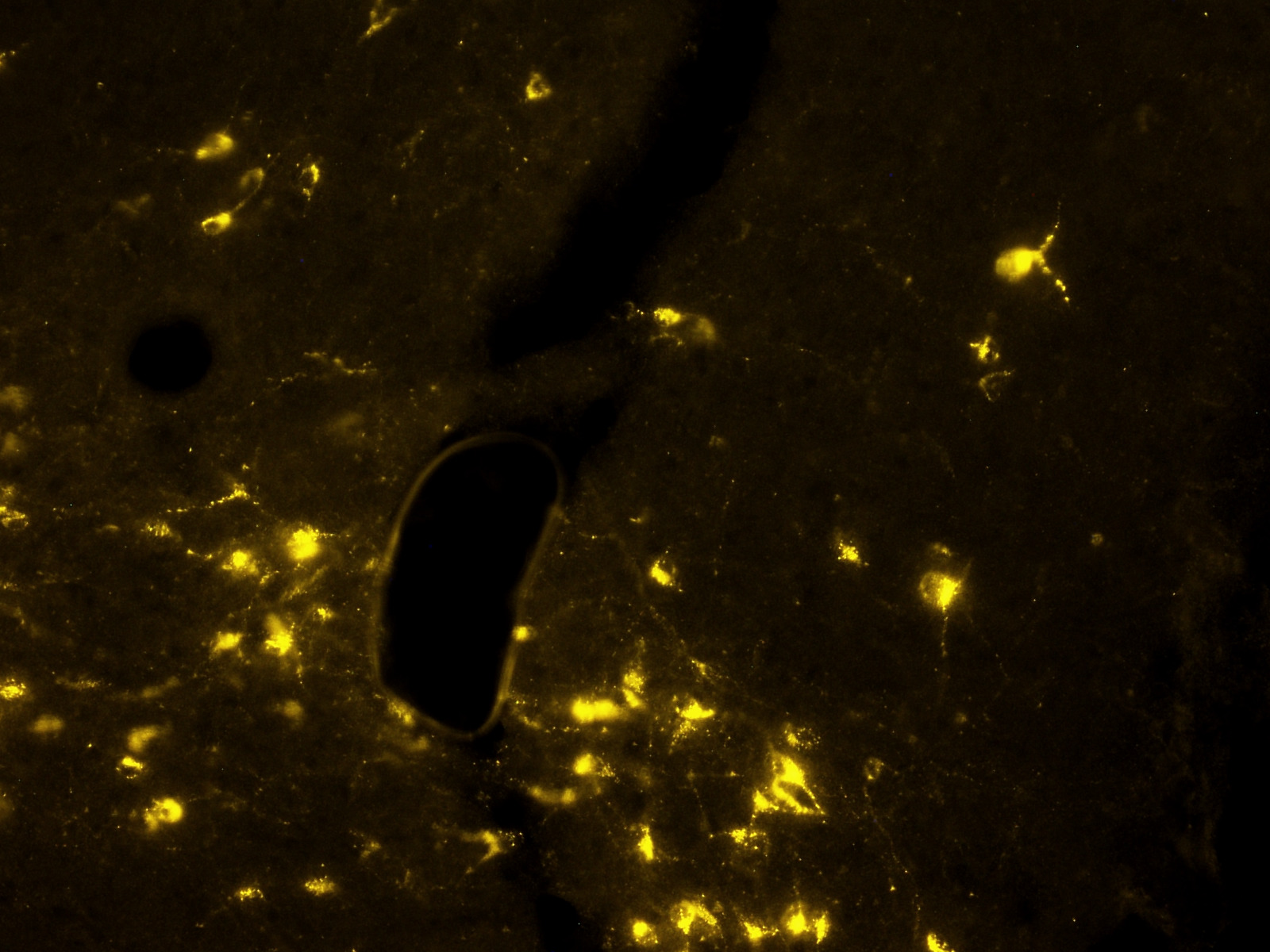}
\includegraphics[width=0.3\textwidth]{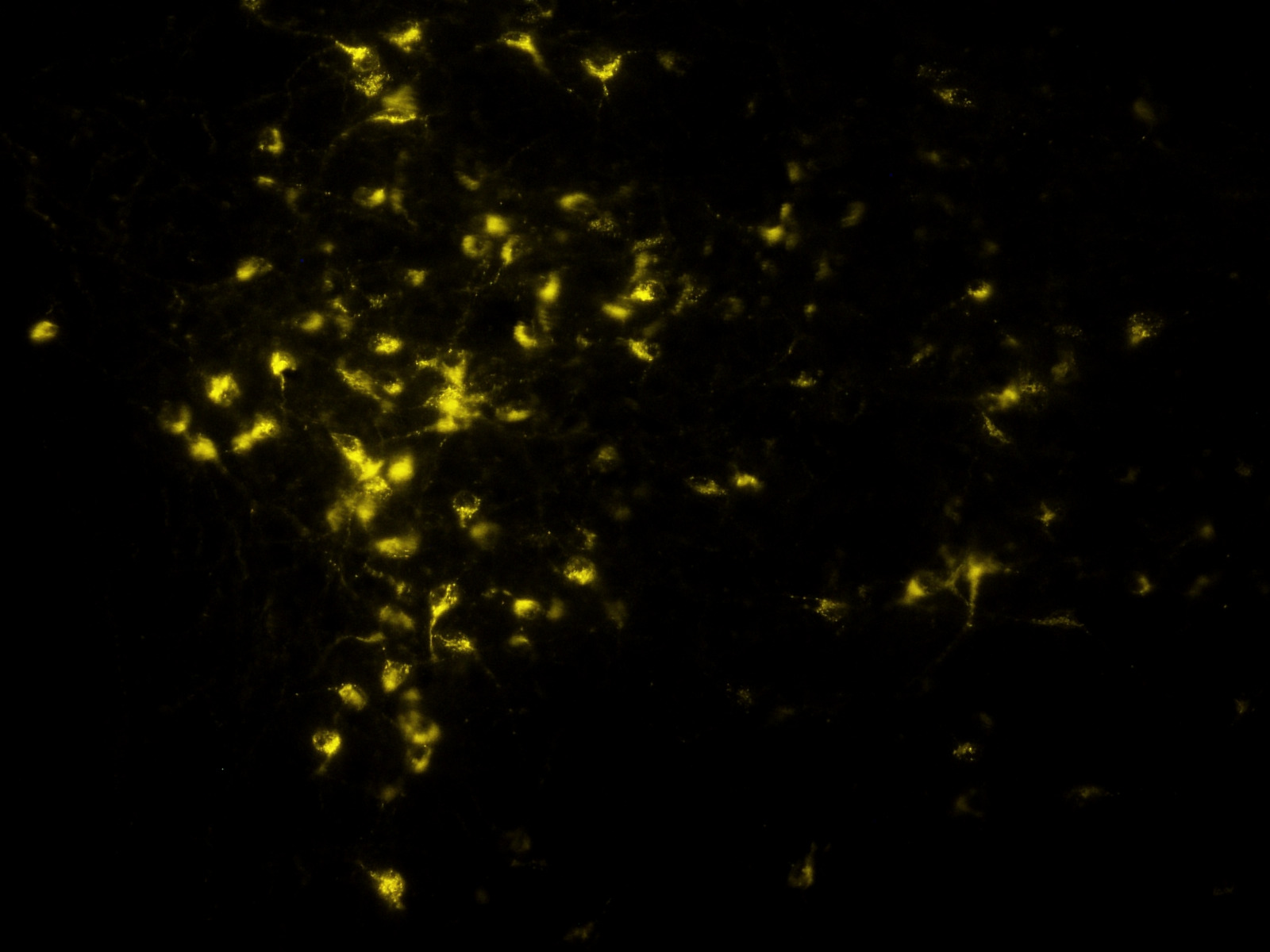}
\includegraphics[width=0.3\textwidth]{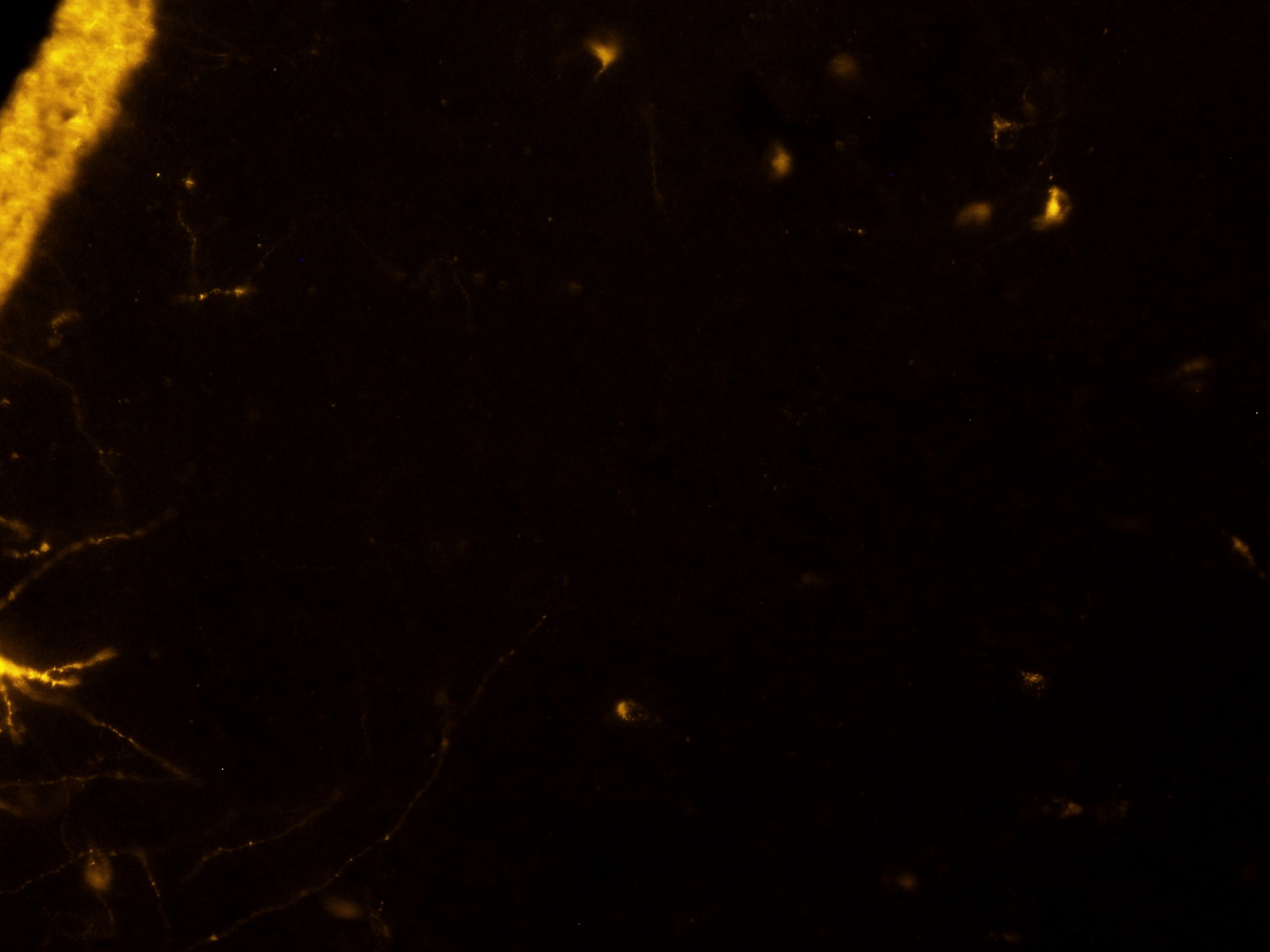}
}
\caption{Sample data \cite{hitrec2019neural}. The neuronal cells of interest appear as yellow spots over a mostly dark background. They exhibit a large variety of shapes, brightness and intensity, which makes them hard to distinguish from artifacts and similar biological structures that are not of interest.} \label{procstructfig1}
\end{figure}

The data \cite{hitrec2019neural} consist of a total of 273 microscopic fluorescence high-resolution pictures (1600$\times$1200) of mice brain slices (Fig. \ref{procstructfig1}). The images represent neurons stained with a monosynaptic retrograde tracer (Cholera Toxin b, CTb), as described in \cite{hitrec2019neural}: CTb was surgically injected in a specific brain structure and, considering the whole brain, only those neurons projecting to that structure (i.e. the injection site) were stained.

\section{Challenges}
\label{sec:challenges}
Although many efforts were made to stabilize the acquisition procedure, the images present some variability in brightness, coloration and other parameters affecting their appearance. Also, the cells themselves exhibit a certain variety of coloration due to the intrinsic variation of the fluorescent emission proprieties. 
Moreover, the substructures of interest have a fluid nature and this implies that the shape of the objects to identify may change significantly, making discriminating between stained cells and background even harder. Finally, the presence of artifacts, biological structures and non-marked cells that look very similar to the activated ones confuses the recognition task. All of these factors 
complicate further the model evaluation given the subjectivity of some borderline cases.

Another source of complexity is due to the wide shift in the number of target cells from image to image. For instance, we have pictures with no stained cells and others showing dozens of cells clumping together. 
In the former case, the model needs high precision in order to prevent false positives. The latter
requires high recall instead, since considering two or more touching neurons only once produces false negatives. 

\section{Method}
\label{sec:method}

\begin{figure}
\centerline{
\includegraphics[width=0.35\textwidth]{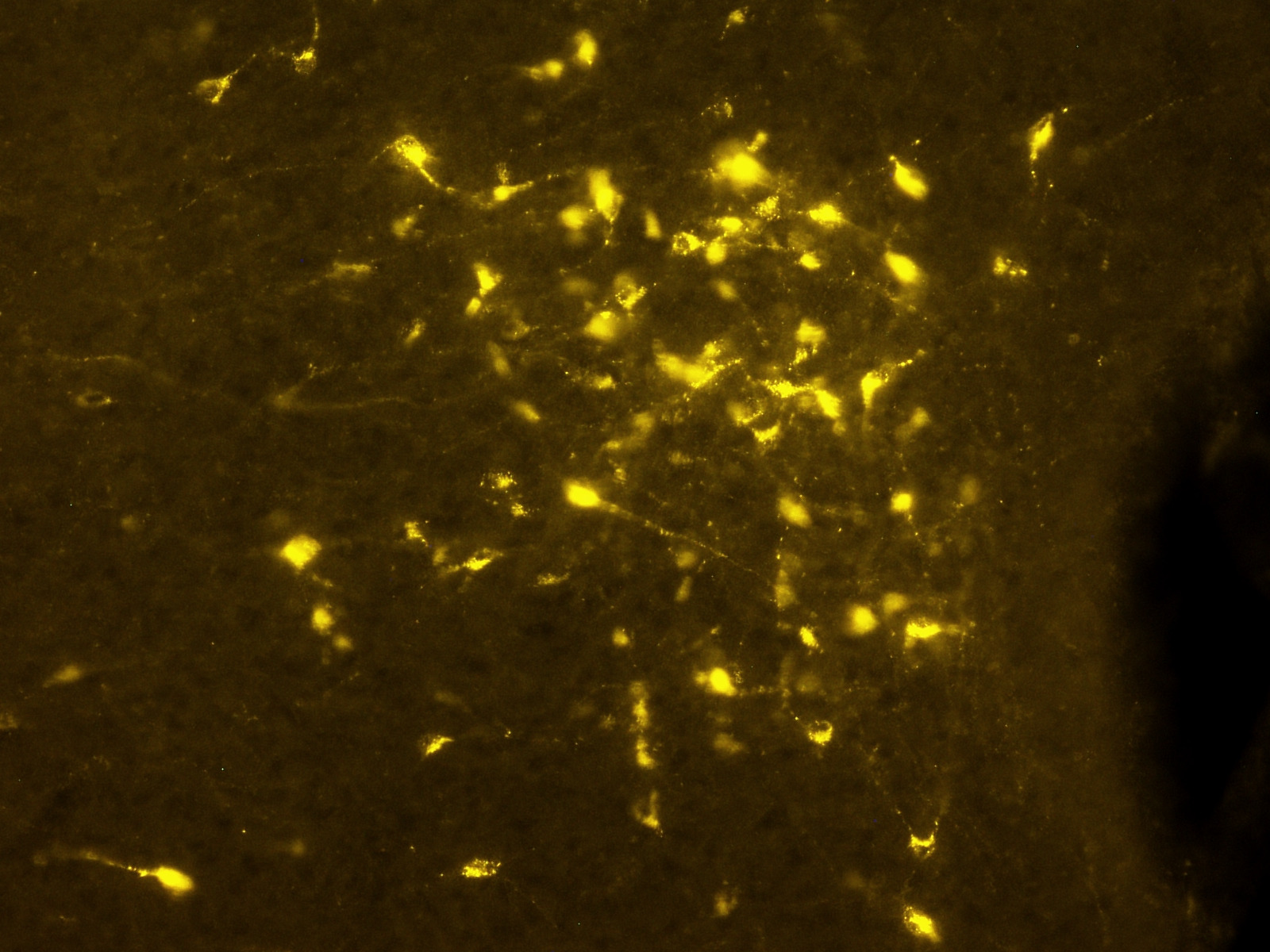}
\includegraphics[width=0.35\textwidth]{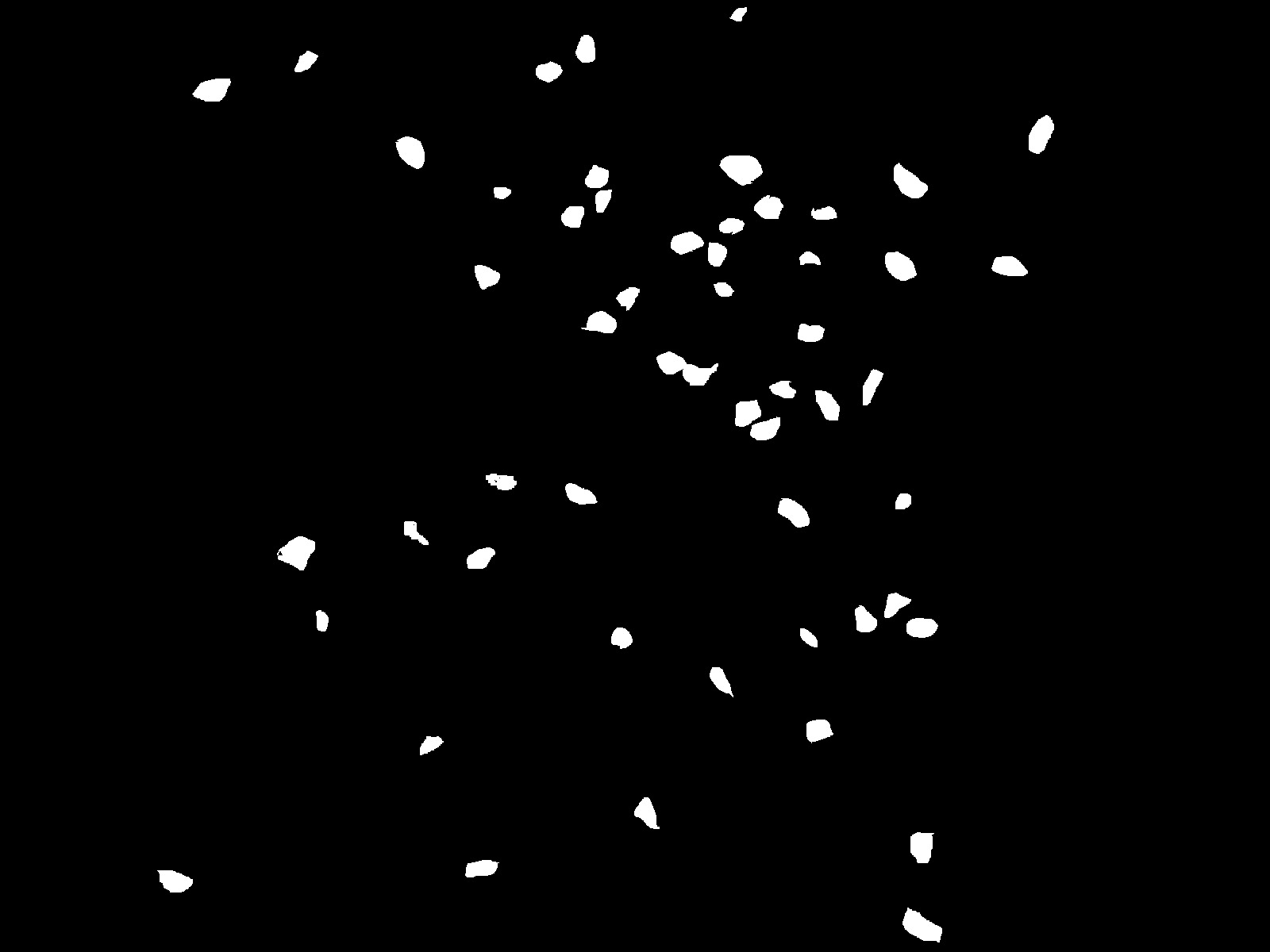}
}
\vspace{0.1cm}
\centerline{
\includegraphics[width=0.35\textwidth]{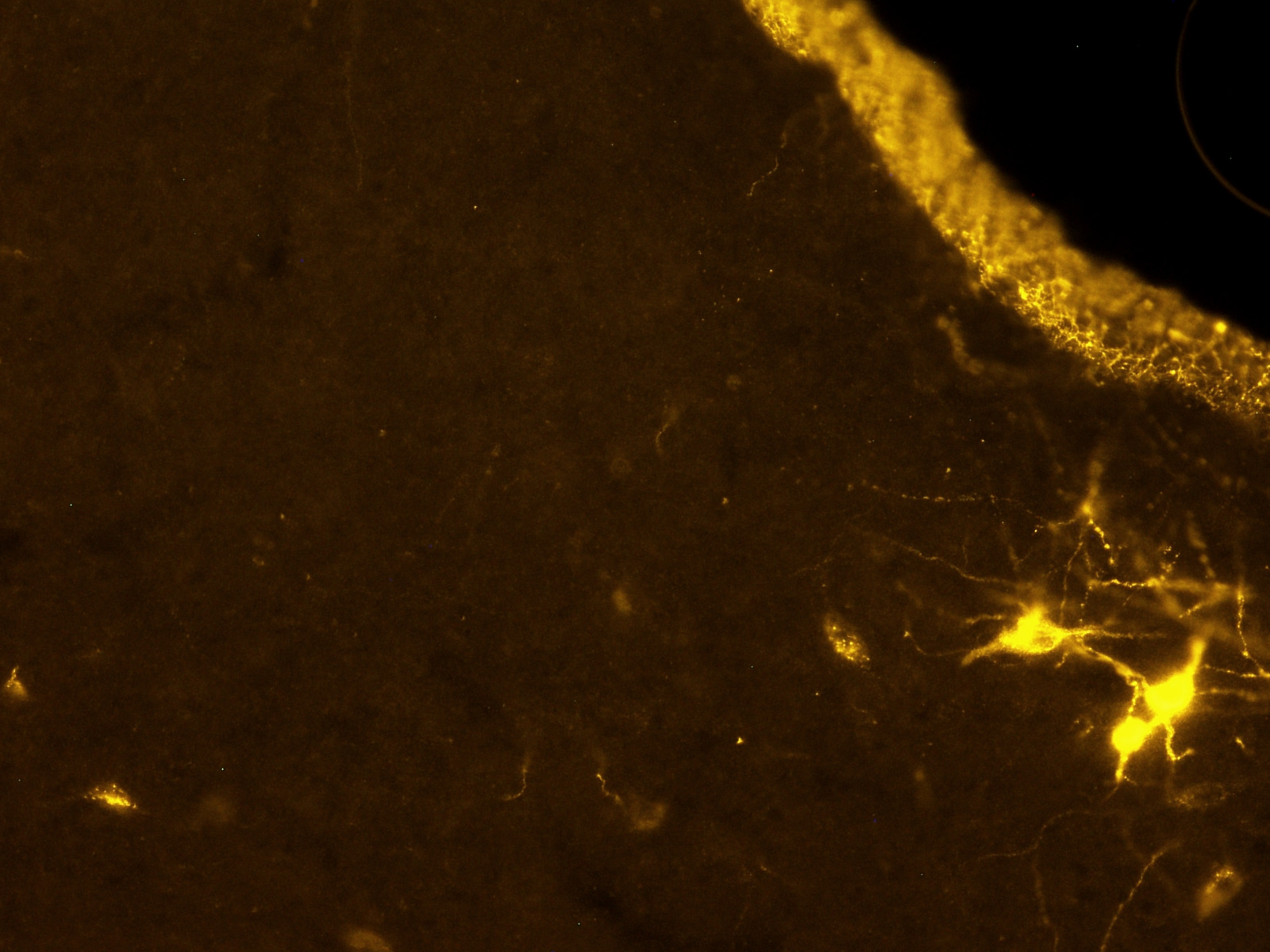}
\includegraphics[width=0.35\textwidth]{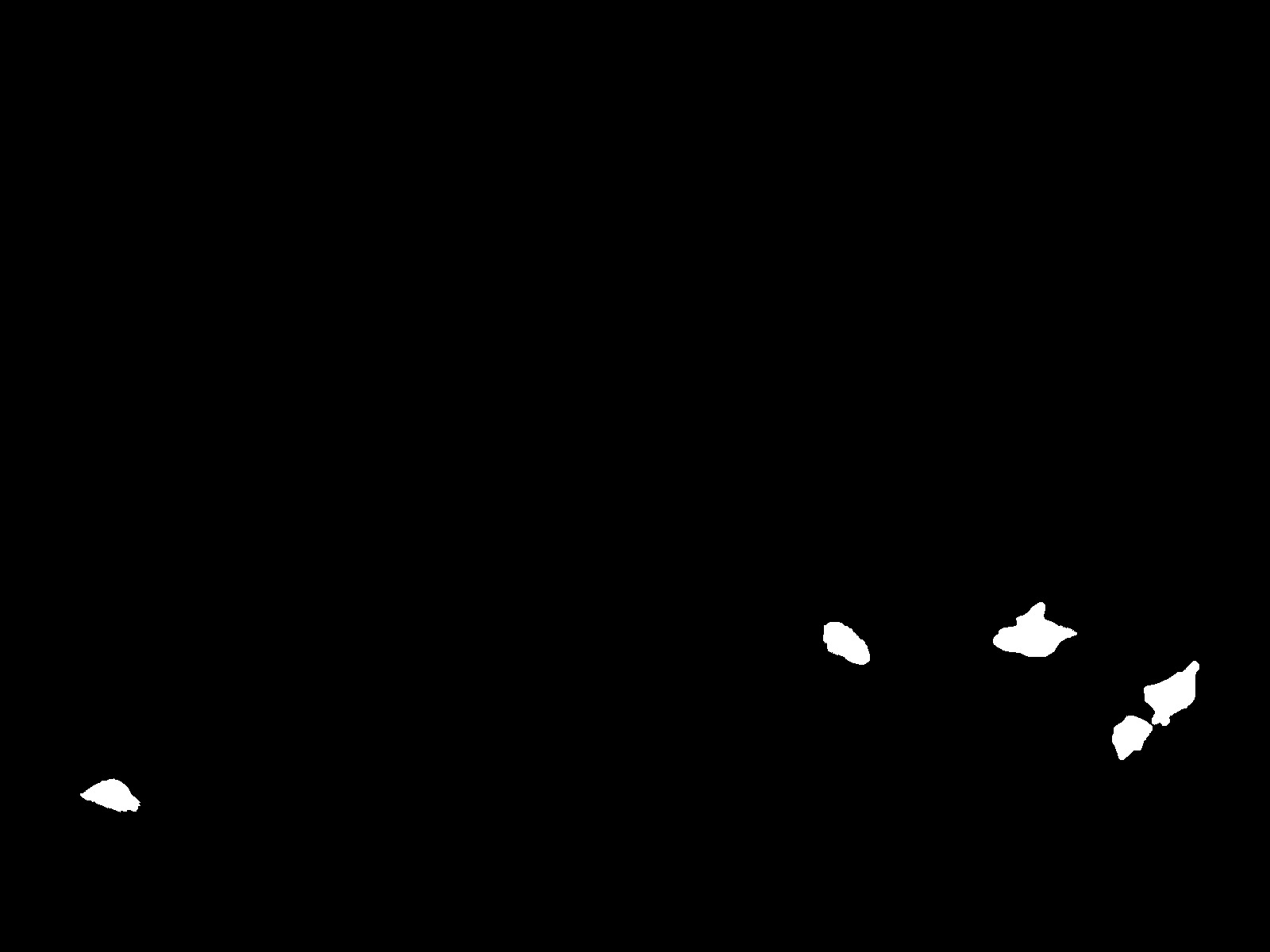}
}
\caption{Images and corresponding target masks.} \label{procstructfig2}
\end{figure}

Training a model for segmentation requires binary masks (\textit{ground-truth}) as targets. In these images, all the cells to segment are represented by white neuronal-shaped objects, while the background is represented with black pixels (Fig. \ref{procstructfig2}).

\subsection{Preprocessing and augmentation}
\label{subcec:preprocessing_augmentation}
Obtaining target masks is usually very demanding so we resorted to an automatic procedure to speed up the labeling. 
In particular, we started from a large subset composed of 252 images and applied common pre-processing filters to remove noise. The cleaned images were then subjected to a thresholding operation based on automatic histogram shape-based methods. 
The goal was to obtain a loose selection of the objects that may seem good candidates to be labeled as neuronal cells. After that, acquainted operators reviewed the results to discard the false positives introduced with the previous procedure, taking care of excluding irrelevant artifacts and tricky biological structures.
The remaining images, were segmented manually by domain experts. We included significant pictures with peculiar traits in the latter set, so to have highly reliable masks for the hardest examples.
Specifically, we selected images with a bigger number of cells, possibly clumping together, and other samples presenting artifacts and/or biological structures that are harder to classify correctly. 
These images were also augmented with a greater factor to make sure they were not under-represented in the training set.

Before the augmentation step, we randomly selected 60 images to insert in the test set, while the remaining pictures were divided into twelve 512x512 partially overlapping crops.
In this way it was possible to handle the same augmentation steps, but with more flexibility. In practice, crops containing some tricky structures were either harder to segment for the network or underrepresented in the sample. By increasing the augmentation factor of these patches, the model improved its performance, especially on artifacts and biological structures that would otherwise present a more troublesome pixel-wise classification.

The augmentation step consisted of both standard manipulations -- like rotations, addition of Gaussian noise, variation of brightness -- and less known methods as the elastic transformation described in \cite{elastic_tranformation}. As a result, the model was trained on a total of nearly 16000 images (70\% for training and 30\% for validation).


\subsection{Weighted maps}
\label{sec:weights_map}

\begin{wrapfigure}{O}{0.55\textwidth}
\centerline{
\includegraphics[width=0.55\textwidth]{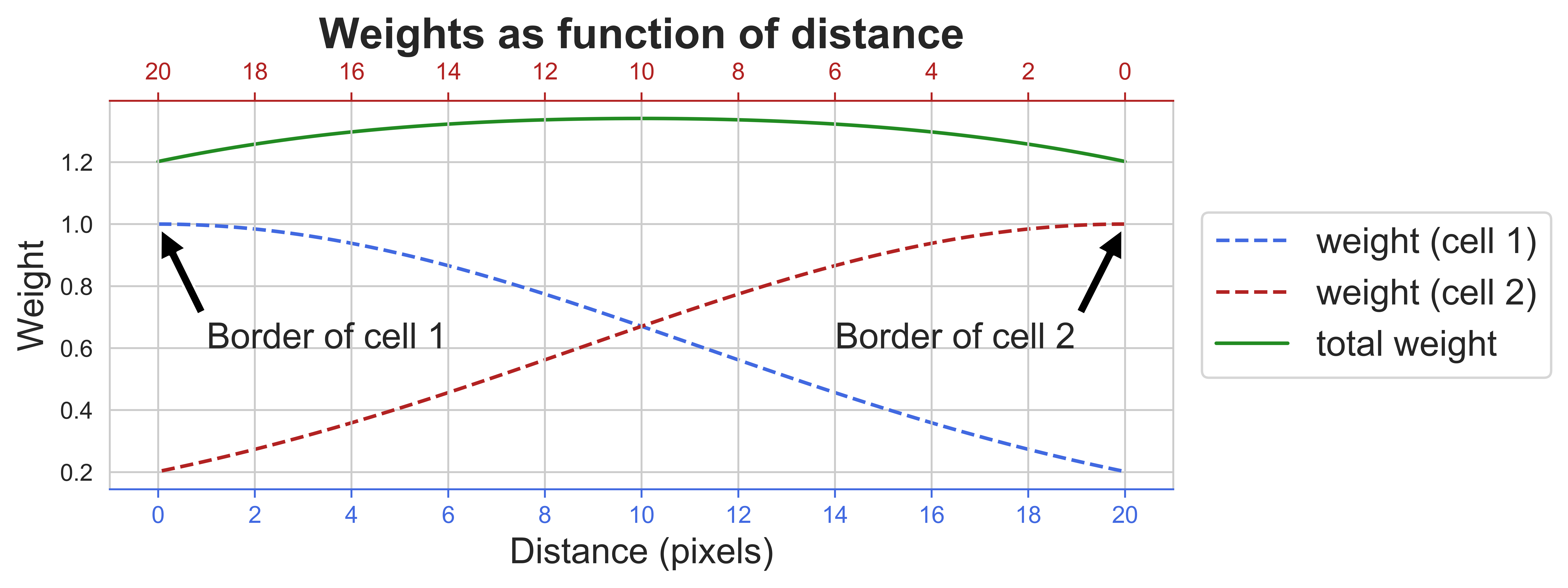}}
\caption{Weight factors of background pixels between clumping cells according to (\ref{weight_formula}). The dashed curves depict the weights generated by single cells as a function of the distance from their borders, respectively cell 1 on the left (blue) and cell 2 on the right (red). The green line illustrates the final weight obtained by adding individual contributions pixel-wise. } 
\label{weight_calculation}
\end{wrapfigure}

One of the most difficult aspects to deal with during the inference is related to cell overcrowding. In this case, it is crucial to avoid counting multiple touching cells only once. This could happen if the segmentation output of the model presents two or more connected objects.
In order to improve cell separation, Ronneberger et al. \cite{unet} suggested leveraging a weighted map that penalizes more the errors on the borders of touching cells.
Here a further development of this idea was considered, increasing the penalization error also with the number of clumped cells, so that the penalization is bigger as the number and/or the proximity of cells increases. To do that we followed these steps for each weighted map:

\begin{enumerate}

    \item initialize a completely black image (all pixels with zero value) for the full weighted maps;
    \item initialize a completely black image for the weighted map generated by the $i$-th item in the target mask;
    \item add the $i$-th object to the map created at step 2);
    \item compute the euclidean distance between each pixel of the current map and the closest pixel belonging to the border of the cell;
    \item compute the weight attached to each pixel of the map according to a decreasing exponential function:
        \begin{align}
        \text{weight} = e^{\frac{-d^{2}}{2\sigma^{2}}}
        \end{align}
        \label{weight_formula}
        where $d$ is the distance computed at step 4) and $\sigma$ is a customizable parameter set to 25 (average cell radius);
    \item sum the resulting current map to the full weighted map, as illustrated in Fig. \ref{weight_calculation};
    \item repeat steps 2) to 6) for each item in the target mask;

\end{enumerate}

This procedure generates weights that decrease as we move away from the borders of each cell. However, the contributions coming from single items are combined additively, so that the full weighted map presents higher values where more cells are close together.

\begin{figure}[b]
\centerline{\includegraphics[ width=0.30\textwidth]{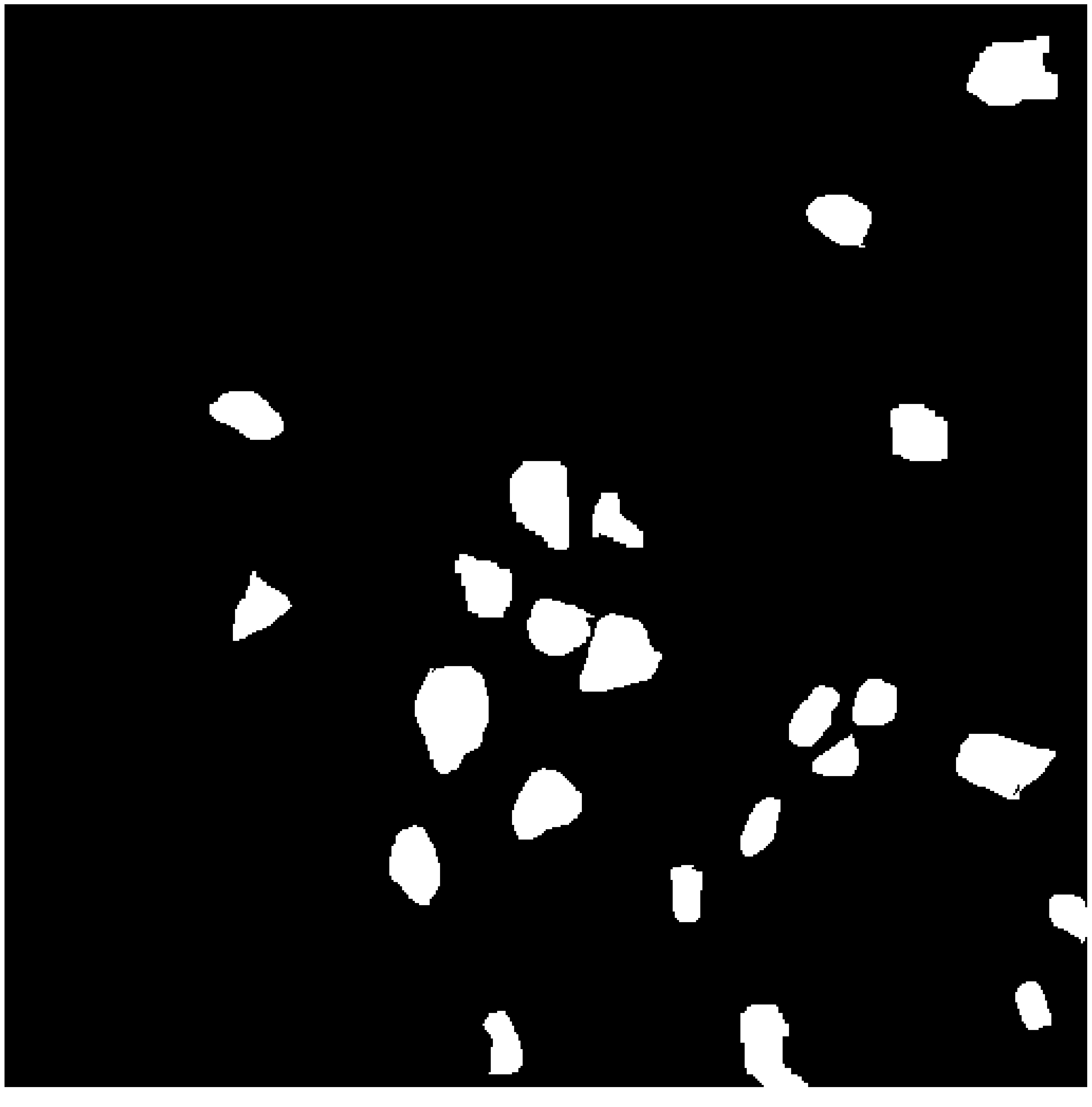}
\includegraphics[trim=0 0.008in 0 0, width=.335\textwidth]{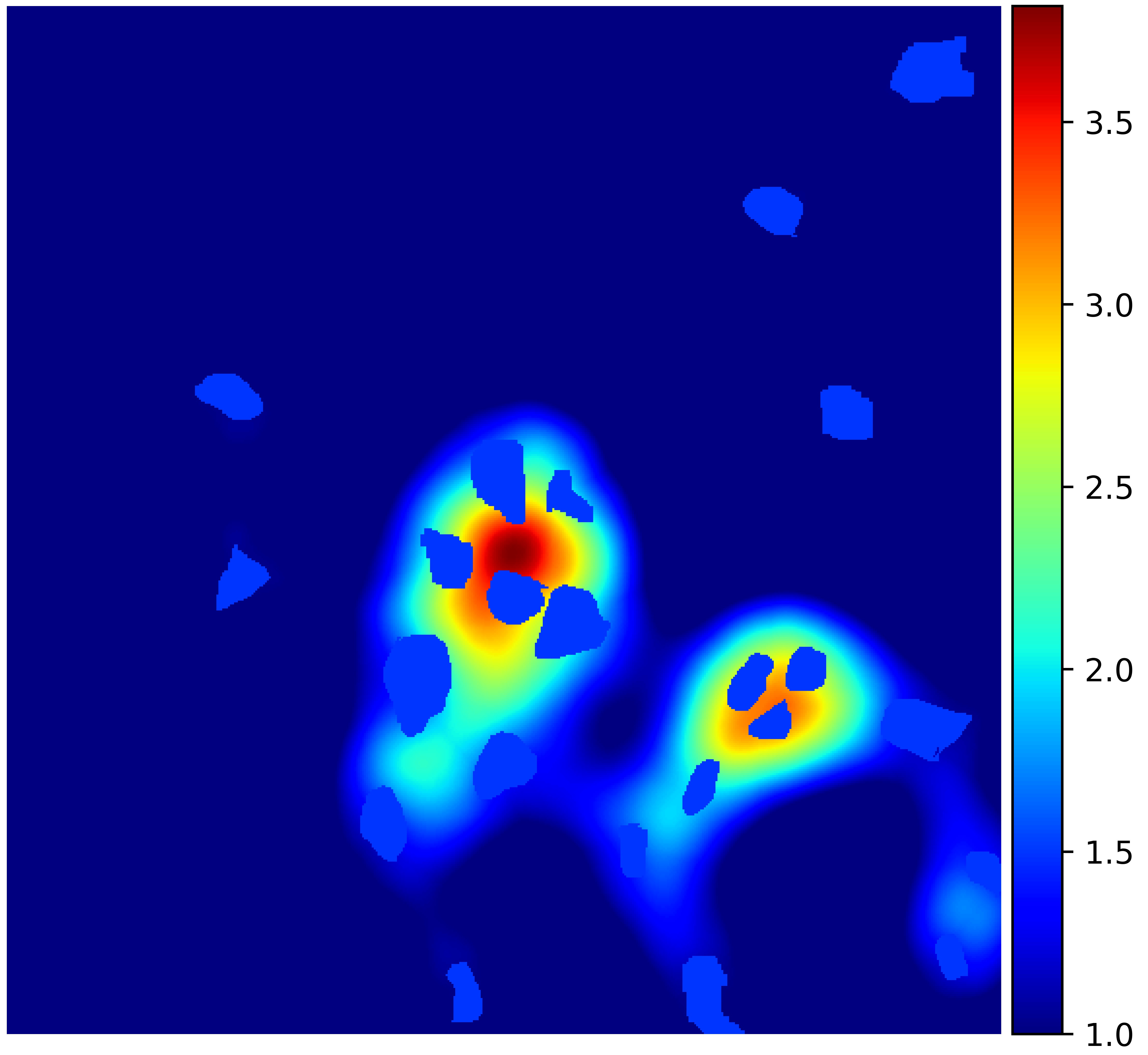}}
\caption{Target mask and corresponding weighted map.} \label{procstructfig3}
\end{figure}

In addition, different weights on both cells and background are applied (1 and 1.5 respectively) to handle the unbalance of the two classes. A weighted map obtained with this implementation is showed in Fig. \ref{procstructfig3}. 

\subsection{Post-processing} \label{sec:post_processing}

A post-processing step is performed to model output after binarization. Specifically, an automated cleaning procedure is applied to remove small noisy components and fill the holes inside the detected cells. After that, the watershed algorithm is employed and its parameters are set based on the average cell size.  This improves the final outcome by splitting most of the overlapping cells in the raw prediction, thus ensuring a higher accuracy.

\subsection{Model architecture}
\label{model_architecture}
\begin{wrapfigure}[33]{O}{0.55\textwidth}
\centerline{
\includegraphics[width=8.5cm]{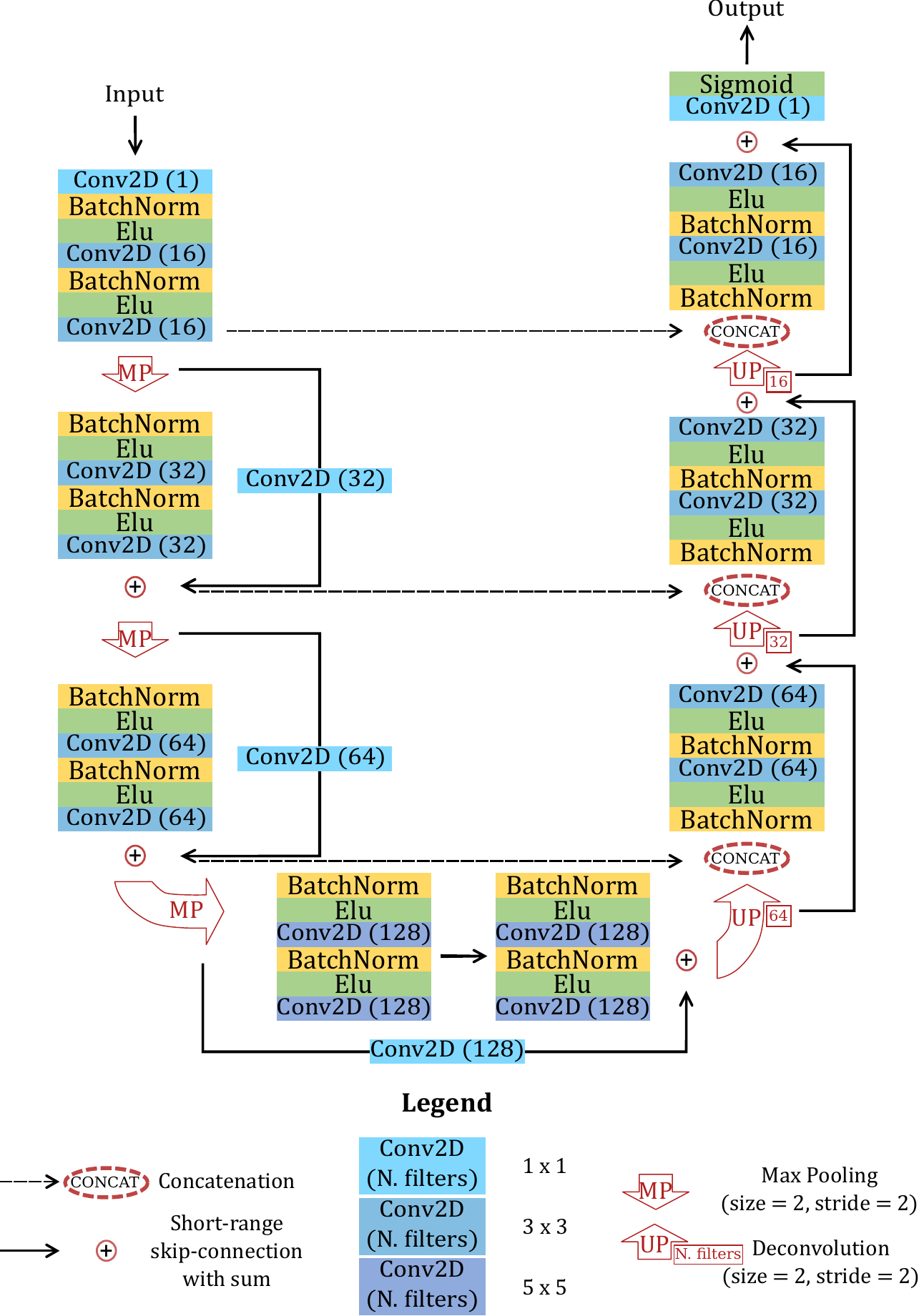}
}
\caption{Scheme of the model. Each box reports an element of the entire architecture (individual description in the legend). The shortcut-connections along the encoding-path are supported by a 1$\times$1 convolution to enable the final sum before the max-pooling operation.} \label{procstructfig4}
\end{wrapfigure}

The model used in this work is strongly inspired by the UNet architecture. As usual, the encoding part is intended to capture what the objects are, while the decoding path refines the localization of the detected items. In addition, we leveraged residual units that facilitate the training and allow a better information propagation as discussed in \cite{resnet}. Specifically, we replaced the plain neural units used in the UNet architecture with residual blocks, as proposed by He et al. \cite{residual_units}. These blocks included both short-range skip-connections (also called shortcut-connections), that enable to learn an identity mapping, and two sequences of batchnormalization, activation and convolution. The final result is similar to the architecture proposed in \cite{deep_resunet} and it is reported in Fig. \ref{procstructfig4}.

On top of that, a 1$\times$1 convolution was used to enable the network to learn an optimal one-dimensional colorspace from the initial three-dimensional space. This is equivalent to applying a grayscale transformation to the RGB input image that is learned and optimized during training.

Another modification with respect to \cite{deep_resunet} was the insertion of an additional residual block at the end of the encoding-path (bottleneck), where we also used bigger size filters (5$\times$5). These adjustments were designed to provide the model with a larger field-of-view, thus fostering a better comprehension of the context surrounding the pixel to classify.
This kind of information can be particularly useful, for example, when cells clump together and pixels on their boundaries have to be segmented. 
Likewise, the analysis of some background structures (Fig. \ref{procstructfig2}, bottom-left image) can be improved by looking at a wider context.

Alternative approaches to increase the field-of-view were also tested. For example, dropping short-range connections \cite{receptive_field} and adopting dilated-convolution \cite{deep_lab} were considered under several configurations, however producing no significant improvements. For this reason, the previous settings were preferred to guarantee a higher resolution in the feature maps and more stability during training. 


Finally, the above architecture was compared with other proposals during the prototyping phase. In particular, both Inception \cite{szegedy2016rethinking, inception_code} and a VGG \cite{Vgg} pre-trained on ImageNet  were tried as encoding module. Nonetheless, these variations did not contribute to improving the validation loss.
We justify these results in two ways. Firstly, the number of parameters increased, thus generating convergence issues. Secondly, the pre-trained weights fitted poorly for our specific task.
In conclusion, the architecture illustrated in Fig. \ref{procstructfig4} produced the best evaluation metrics and it was also more reliable and stable during the learning phase.

\subsection{Model training}
\label{sec:model_training}
The model was implemented in Tensorflow through Keras and the training was performed by exploiting 4 V100 GPUs provided by the \textit{Centro Nazionale Analisi Fotogrammi} (CNAF) computing center in Bologna. The loss used to train the model was the \textbf{weighted binary cross-entropy} with weighted maps defined as described in section \ref{sec:weights_map}.
During the training, the images (512$\times$512 crops) were fed in the network batch-by-batch using a generator. In this phase, a normalization occurred whereby each pixel intensity was divided by 255 over all of the RGB channels. The train started with a learning-rate of 0.006 and a scheduler decreased its current value by 30 \% every time the validation loss did not improve for 4 consecutive epochs. The Adam optimizer was used, showing a better stability over the stochastic gradient descent in almost all the trials. The model converged in less than 50 epochs.
\section{Results}
\label{sec:results}
\begin{figure}
\centerline{\includegraphics[width=0.83\textwidth]{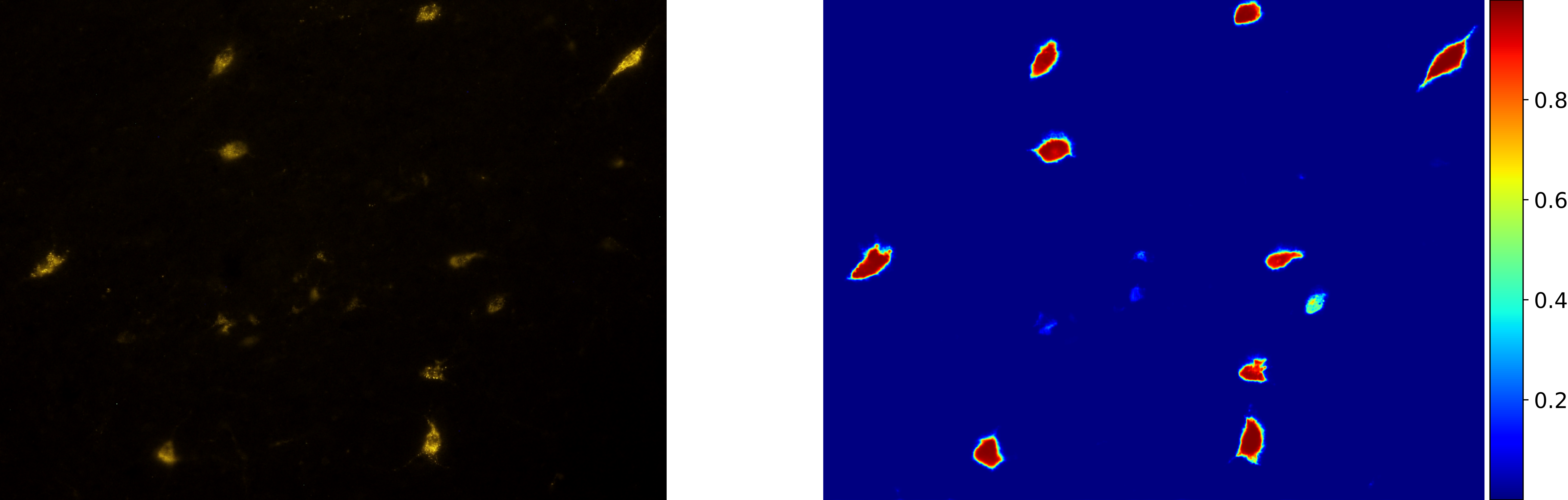}}
\centerline{\includegraphics[width=0.83\textwidth]{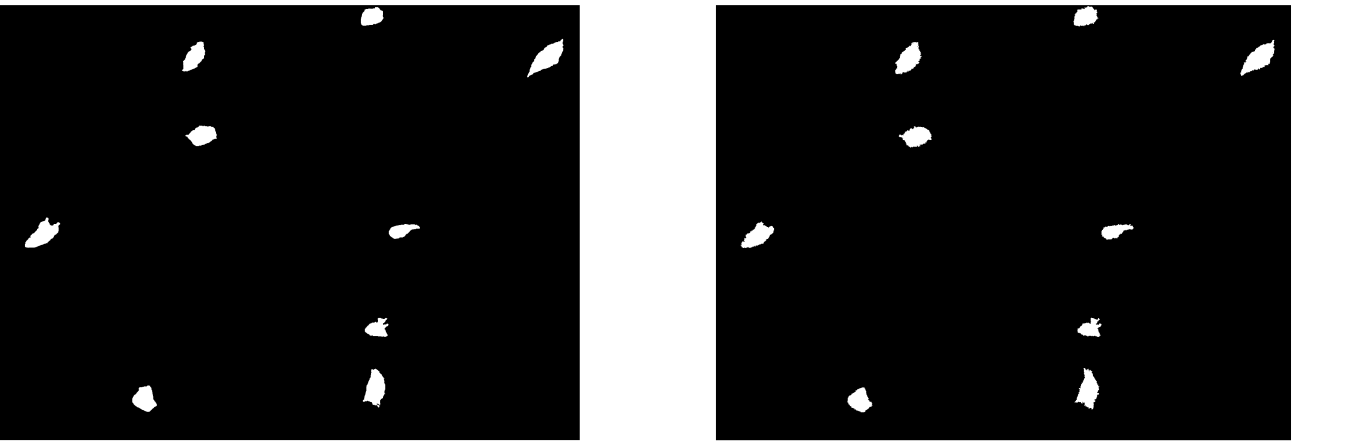}}
\caption{Raw output and post-processing. In the top row we have the input image (left) and the model's raw output (right). Below, the predicted mask after post-processing (left) is compared with the target (right). }
\label{procstructfig5}
\end{figure}

The fully-convolutional architecture allows predicting on full-size images (1200$\times$1600) despite 512$\times$512 crops were used during the training. The final output of the model is a probability map (or heatmap), in which each pixel represents the probability of belonging to a cell (Fig. \ref{procstructfig5}, top-right plot). The higher the value, the higher is the confidence in classifying that pixel as belonging to a cell. A thresholding operation was then applied on the heatmap so to obtain a binary mask where the detected cells are represented by groups of white connected pixels. Before actually counting the items, an ad hoc post-processing was performed as described in section \ref{sec:post_processing}.

\begin{wrapfigure}[18]{O}{0.55\textwidth}
\centerline{\includegraphics[width=.55\textwidth]{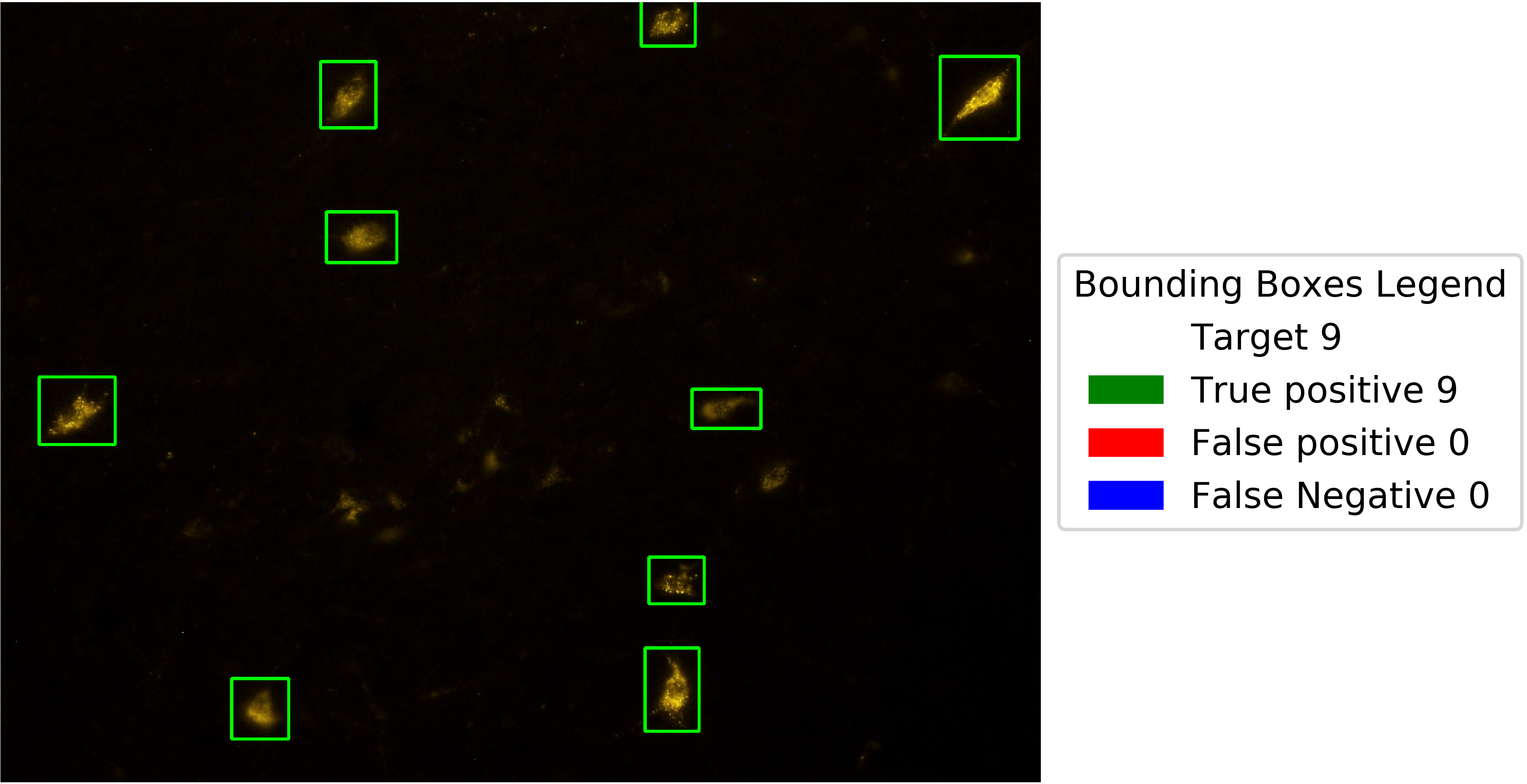}}
\caption{Bounding box visualization. Green bounding boxes represent true positives, while prediction errors are illustrated in red (false positives) or blue (false negatives). In this case a perfect agreement is achieved.} \label{procstructfig6}
\end{wrapfigure}

Looking at Fig. \ref{procstructfig5} it is possible to appreciate the discrimination power of the model. The heatmaps are characterized by a significant spread between the predicted probabilities of pixels classified as cells and the background. This means that the model is very confident of most of its predictions. 
Another very good feature is the capacity to avoid all neurons not clearly stained, as the ones localized near to the center of the top-left image. Actually, that group of cells is present in the heatmap, but with a lower confidence level. Hence, if necessary, these cells can still be detected lowering the threshold. 
A comparison between target and detected objects is reported in Fig.\ref{procstructfig6}. 

\subsection{Model evaluation}
\label{sec:model_evaluation}

In this section we report the results of the training, with a focus on both localization and counting performances on the test set.

As far as the segmentation task, we accomplished good results and achieved a \textit{mean intersection over union} (Mean IOU) of \textbf{0.74}. The model was able to recognize different cell shapes and draw accurate boundaries, also managing to separate clumping items in most cases.
However, we are not really interested in an exact pixel-wise segmentation inasmuch it serves just as an high-level justification of which objects are contributing to the final counts.

For this reason, we evaluated the model in a classification fashion by means of the $\mathbf{F_{1}}$ \textbf{score}. This metric is more suitable for our purpose since it is focused on which are the detected cells rather than how well they are represented. 

\begin{wrapfigure}{O}{0.55\textwidth}
\centerline{
\includegraphics[width=\linewidth]{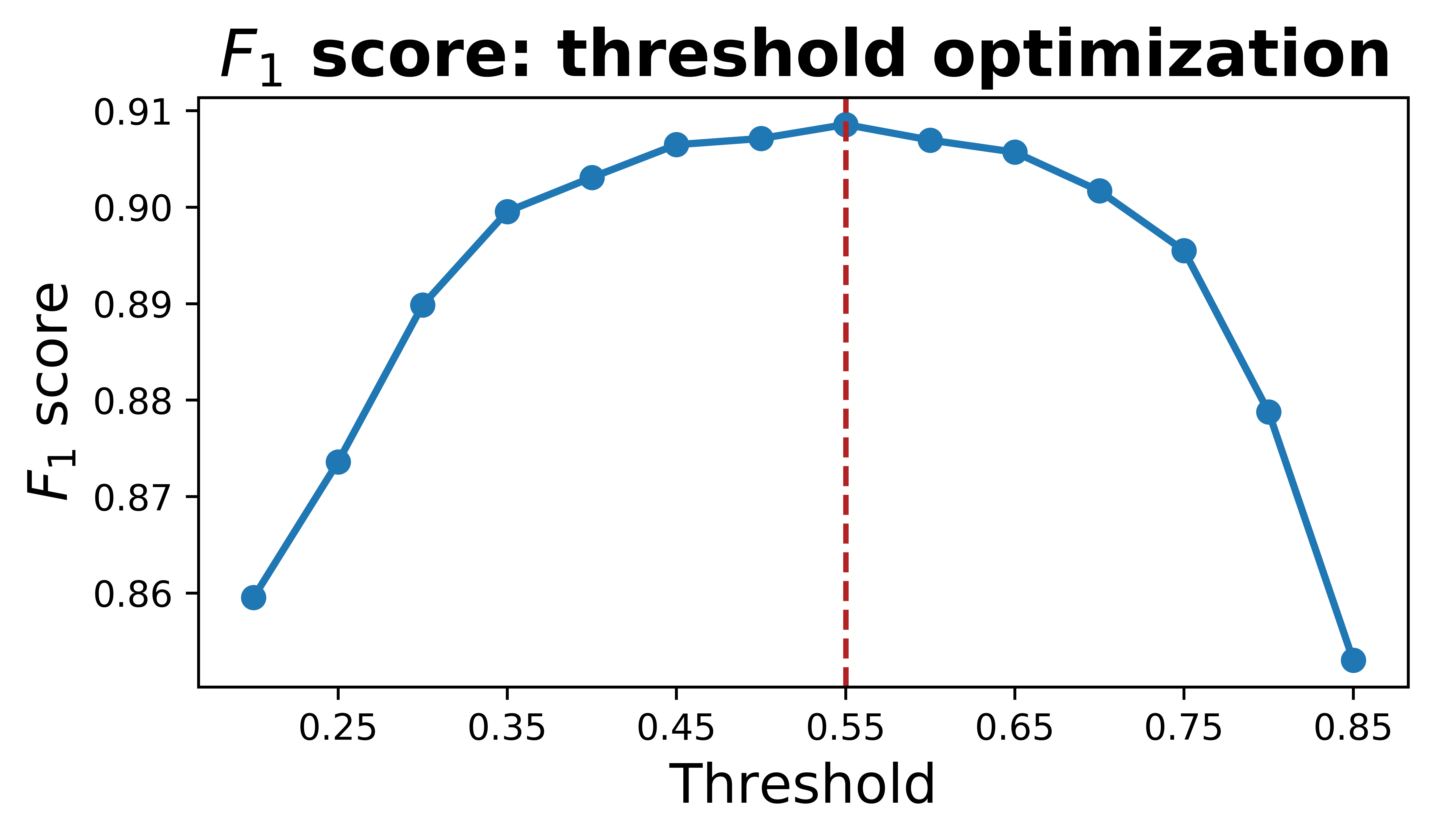}
}
\caption{Threshold optimization. Trend of the $F_{1}$ score computed on the training images as a function of the cutoff for thresholding.} \label{procstructfig7}
\end{wrapfigure}

As a first step, we transformed the raw heatmaps produced by the model into binary masks through thresholding. The optimization of the cutoff was based on the $F_{1}$ score computed on the training set (Fig. \ref{procstructfig7}), resulting in a best threshold of 0.55. 
The optimal value was then adopted to assess the model performance on the 60 test images. In practice, a dedicated algorithm was developed to determine the correctness of model predictions.
Specifically, each target cell was compared to all of the objects in the corresponding predicted mask and uniquely associated with the closest one. If the distance between their centroids was less then a fixed threshold (50 pixels, i.e. average cell diameter), the predicted element was considered as a true positive; a false negative otherwise.
Detected items not associated to any target were considered as false positives instead.
The final result was an $F_{1}$ score of \textbf{0.87} computed on the 60 test images.

\begin{figure}[b]
\centerline{\includegraphics[width=0.8\textwidth]{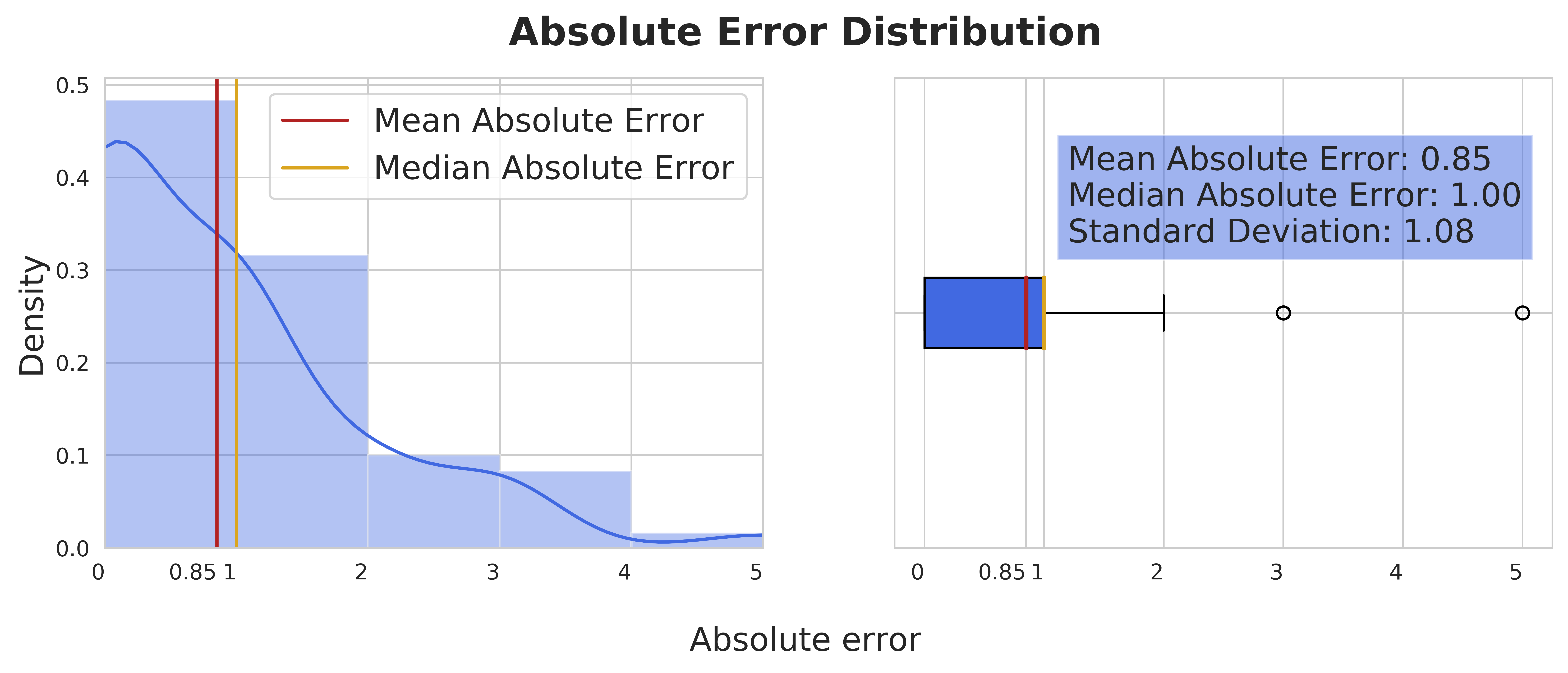}}
\caption{Counting metrics. Boxplot (left) and distribution (right) of the counting absolute error.} \label{procstructfig8}
\end{figure}

The counting performance was analyzed in terms of \textit{mean} and \textit{median absolute difference} (MAE and MedAE respectively) between true and predicted counts. The results are reported in the Fig. \ref{procstructfig8}. 
The model seemed to perform well also on this task, with average and median absolute errors close to 1.
A summary of the results is reported in table \ref{table1}.
\begin{table}[ht]
\begin{center}
{\caption{Performance metrics over the test set. The best cutoff (0.55) was used for thresholding.}\label{table1}}
\begin{tabular}{cccc}
\hline
\rule{0pt}{12pt}
\textbf{$\mathbf{F_1}$ Score} & \textbf{MedAE} & \textbf{MAE} & \textbf{Mean IOU}\\
\hline
0.87 & 1 & 0.85 & 0.74\\
\hline
\end{tabular}
\end{center}
\end{table}
\label{sec:discussion}
\section{Discussion}

\begin{figure}[t]
\centerline{
\includegraphics[width=0.35\textwidth]{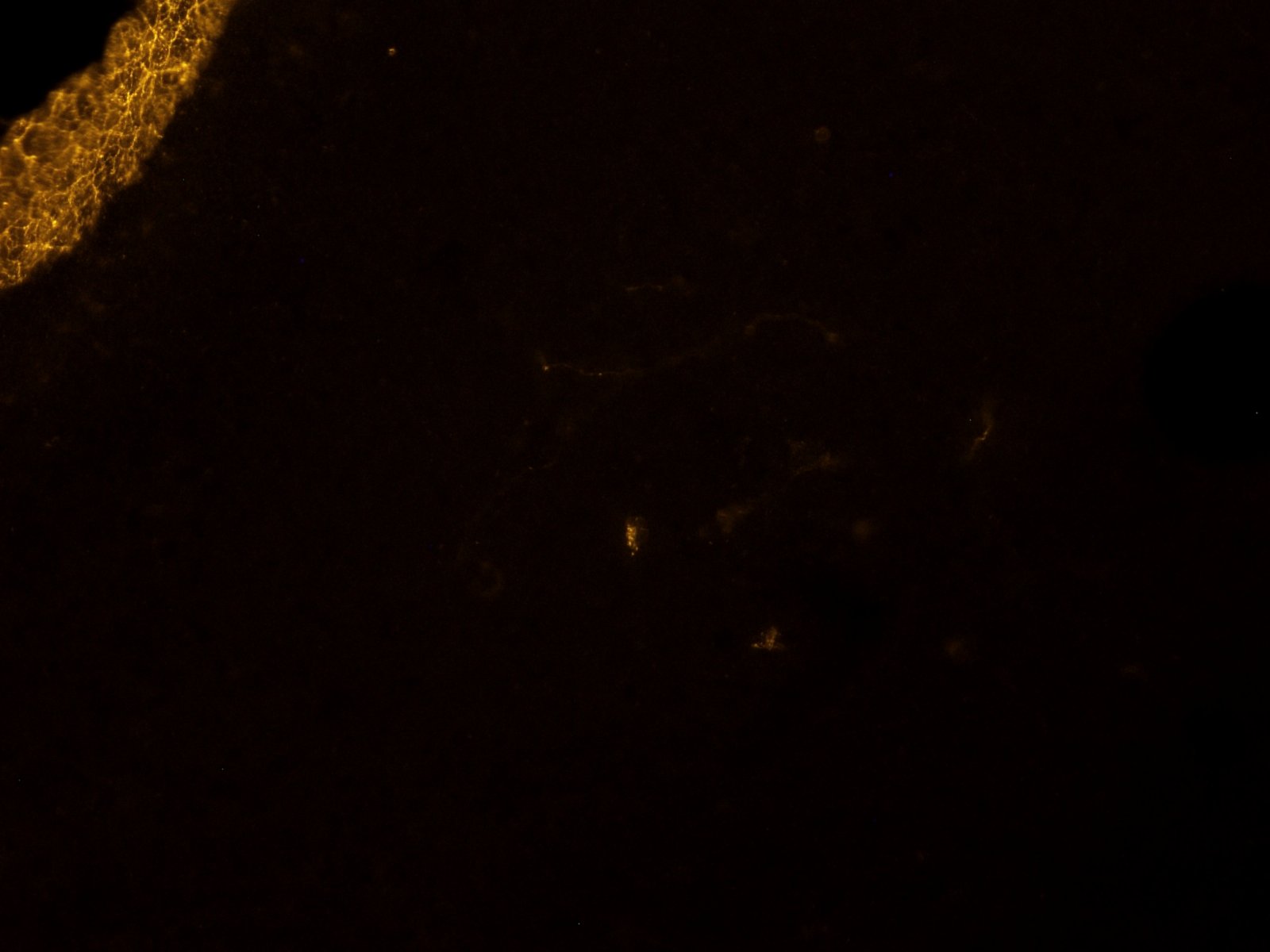}
\includegraphics[width=0.35\textwidth]{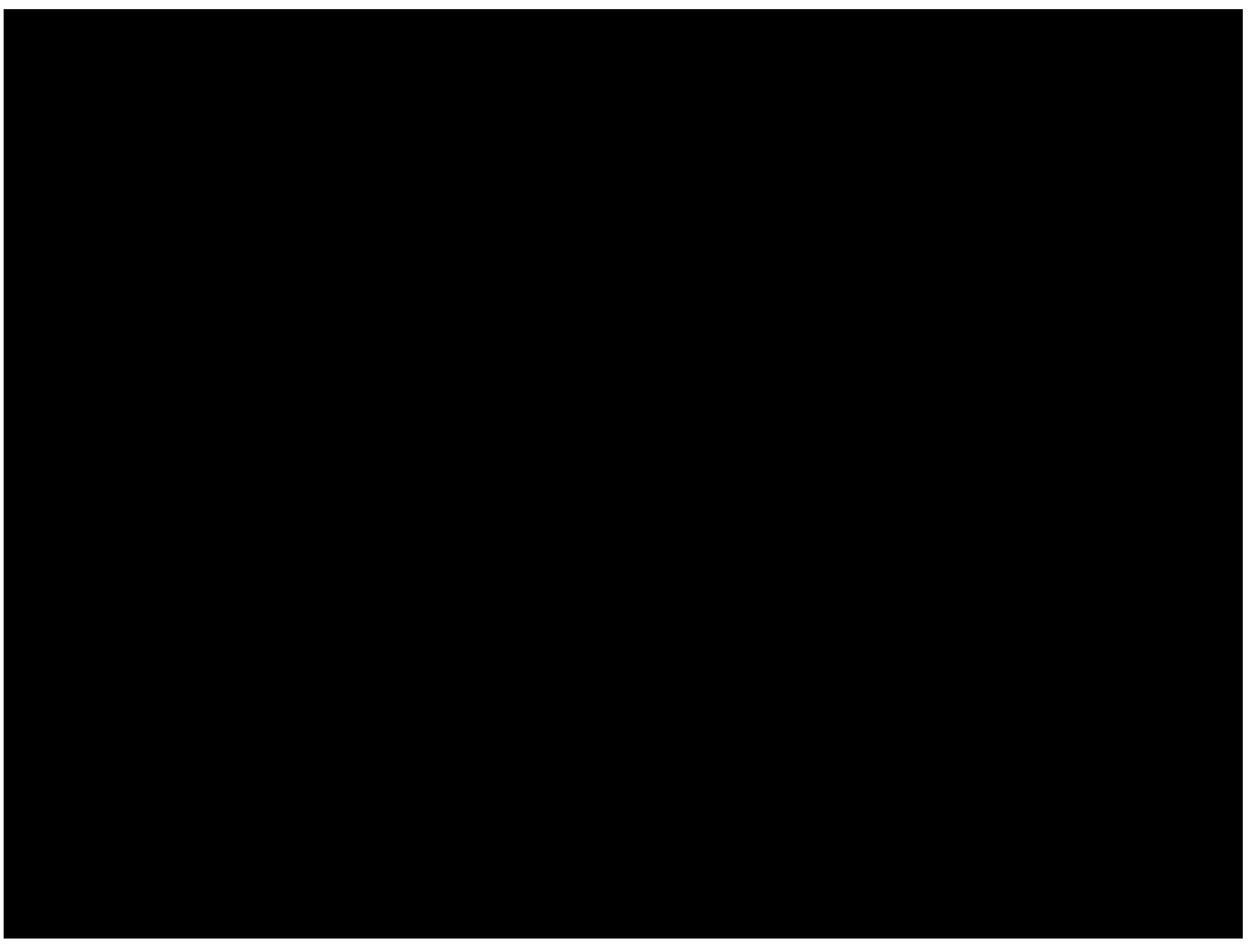}
}
\caption{Model prediction on a test image. Notice the yellow strip was correctly ignored in the output.} \label{procstructfig9}
\end{figure}

Supervised methods use reliable targets to let the model learn from examples. In this work we trained a model starting from masks generated with both standard image processing methods (and corrected a posteriori by experts, if necessary) and hand-made segmentation. The model captured salient features from either samples and showed no bias during the qualitative assessment of the results. 
In our view, this is due to a specific augmentation pipeline targeted to better exploit the crops with relevant characteristics to learn. For instance, looking at Fig. \ref{procstructfig9} we can observe how the yellow strip on the upper-left corner is not erroneously identified as part of a cell (or as multiple cells). On the contrary, it is worth mentioning that the same did not happen without a dedicated oversampling of this kind of biological structures. In fact, the model tended to be fooled by the intense coloration of the pixels in that area since such traits were underrepresented in the original sample, thus leading to false positives. The same behavior is not trivial to obtain with a standard thresholding approach that takes into account only global properties of the image. 
\begin{figure}
\begin{wrapfigure}[28]{O}{0.55\textwidth}
\includegraphics[width=0.53\textwidth]{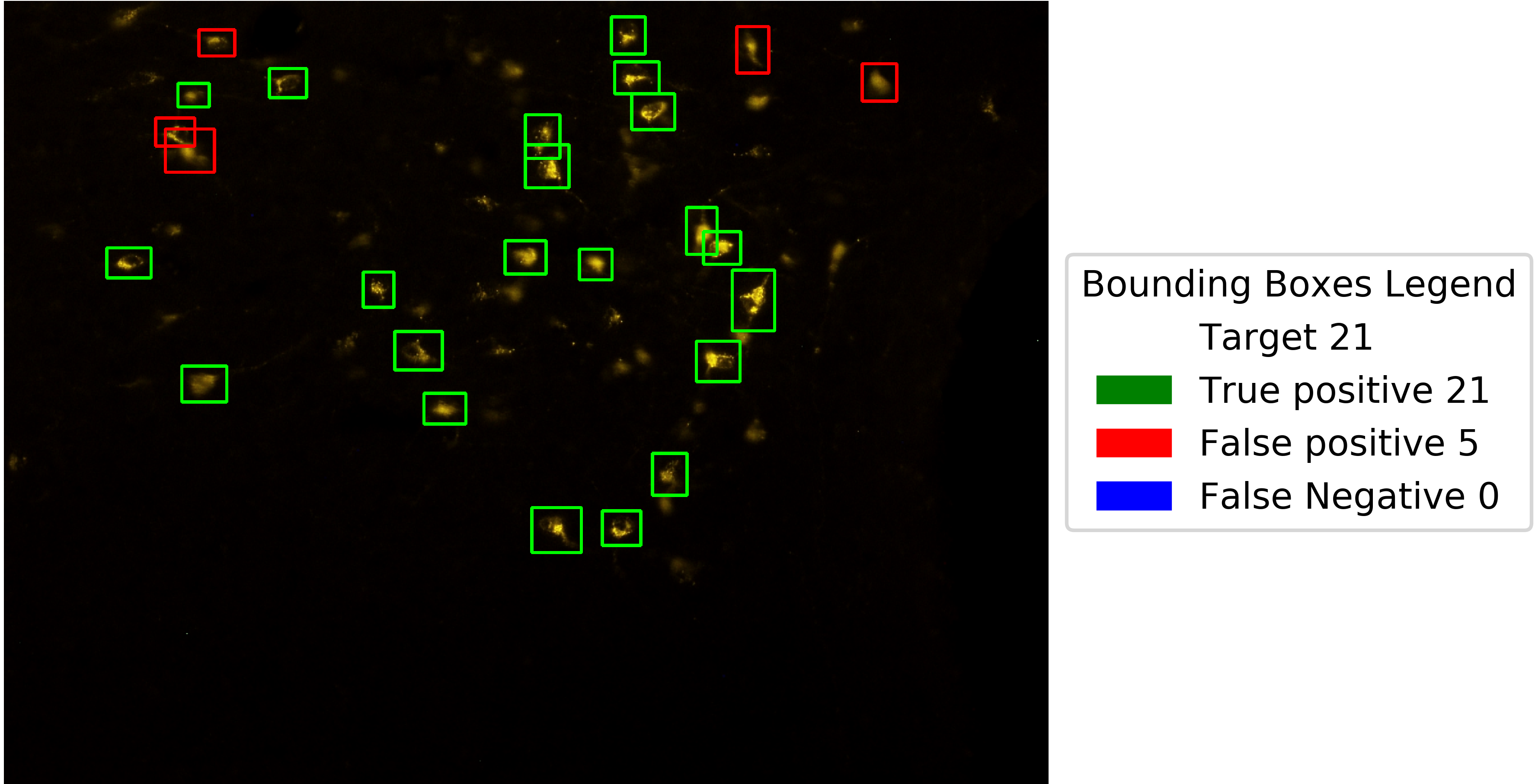}
\vspace{0.1cm}
\includegraphics[width=0.53\textwidth]{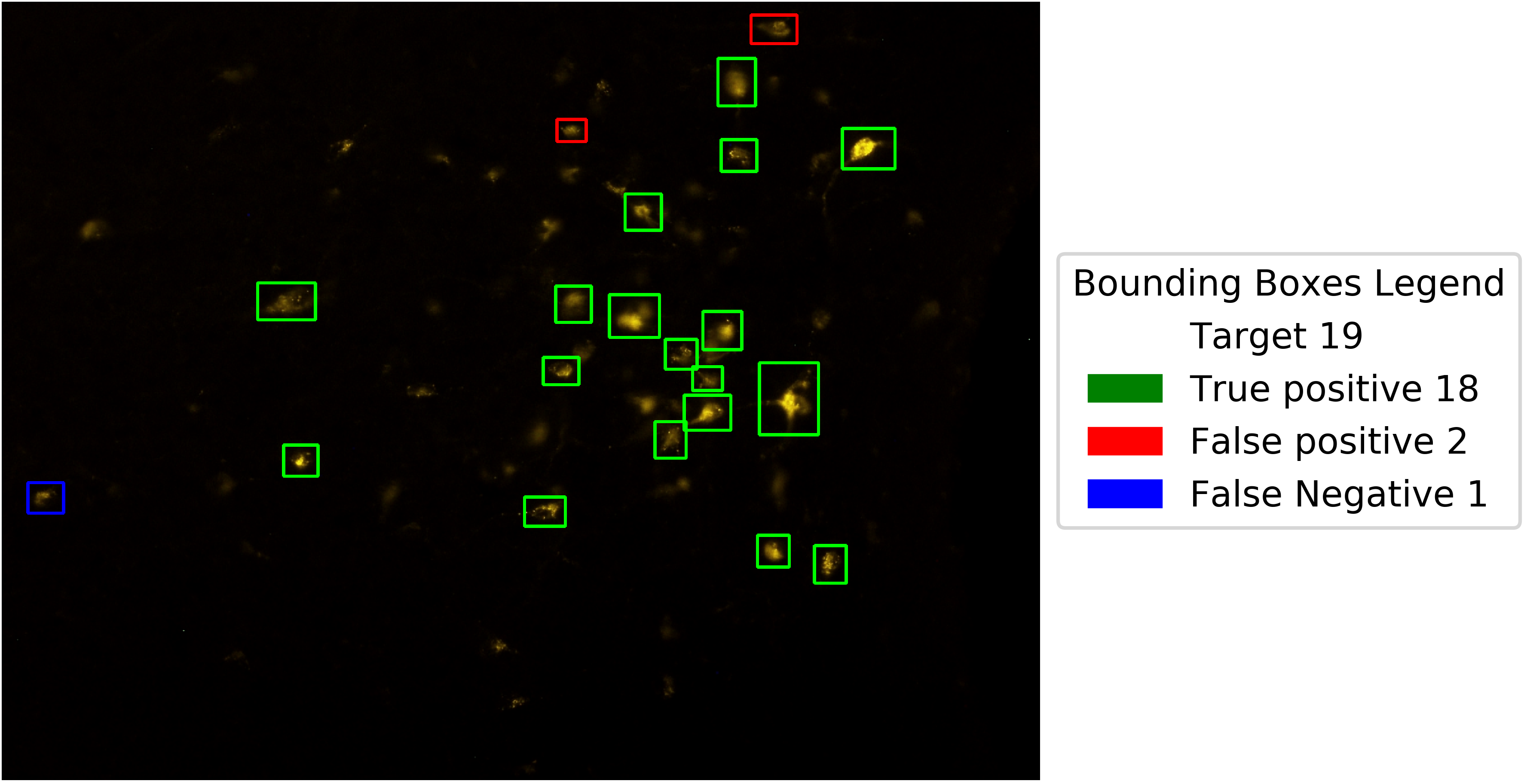}
\caption{Results on test images. The model performs very well in most of the cases. False positives (red) look like target cells. Likewise, the object discarded in the bottom image (blue) is similar to other spots that were not annotated. Thus, even the errors lay within the uncertainty due to operator's interpretation.} \label{procstructfig10}
\end{wrapfigure}
\end{figure}
The weighted masks helped the model extracting features to better contextualize  cells in highly crowded regions. As a consequence, touching cells were effectively detected as distinct items in most of the cases.
Fig. \ref{procstructfig10} illustrates how hard it can be to distinguish false positives from true positives. In general, all the borderline cases bring a certain amount of subjectivity in the counting process, thus making decisions arduous even for experienced operators and hampering an automatic quantitative evaluation of the results.
This difficulty was confirmed through a visual assessment made by qualified annotators, revealing that a considerable part of the model errors lay within the limits of subjective interpretation of borderline cases. 
Nonetheless, the amounts of false positives and false negatives often compensate each other, thus not affecting much the goodness of fit with respect to the counting task.
In conclusion, the global performances of the model were positively evaluated from domain experts.

To obtain these results we also decided to enlarge the field-of-view with respect to the standard UNet architecture, but without significantly increasing the number of parameters to avoid overfitting. 
In this regard, exploring dilated convolution could improve at least the segmentation ability of the model without further increasing the number of parameters. 
However, inserting different filters did not lead to appreciable improvements, instead a lower precision in the segmentation of cells highlighted a difficulty to go beyond the artifacts usually generated from the \textit{atrous-convolution} \cite{gridding}.

As a final remark, notice that the adoption of a fully-convolutional architecture allows applying the model on images whose size is different from the one of the training pictures.
\section{CONCLUSION}
\label{sec:conclusion}

In this work, we trained a fully-convolutional neural network  on fluorescent microscopy images to retrieve the number of stained neuronal cells without human intervention. As a result, we showed that it is possible to design a reliable model focusing on the more problematic samples during the augmentation process. The performances on test images did not highlight systematic failings, neither on particularly low-quality images nor when very challenging examples were presented.

The issue related to overlapping cells could be mitigated adopting weighted maps that also account for the entity of the overcrowding. We finalized all of these improvements giving a special focus to the field-of-view. One possible method to enhance the context information during the pixel-wise classification is to add more convolutional layers and to increase the size of the filters, as we did in the bottleneck of the architecture. 
In conclusion, the final model demonstrated high performance in recognizing cells with diverse shapes and color intensity. Also, a posterior assessment conducted by domain experts revealed a \textit{human-level} ability in the image analysis, as testified by the possibility to interpret erroneous predictions as borderline cases.

\subsection{Future Work}
\label{sec:future_work}

The most natural follow-up of this work is testing the model on similar use cases. The study underpinning our analysis, in fact, involved the injection of different markers to the same brain samples, so to highlight multiple biological structures of interest. As a consequence, a possible development may be to extend the model to account also for similar cells differing for color or shape.
One way to do that would be to include these new pictures in the analysis and train the model on multiple colorations. Alternatively, a domain adaptation approach could be investigated. In this case, the training performed here could be exploited as a surrogate task for speeding up and improving the learning process of the new cells. 

A further development in the perspective of a complete automation is deploying this model on an online platform. This solution would provide an Application as a Service (AaaS) that would foster the studies in this and similar areas. 
In this way, researchers could access the tool on-demand, upload input images and receive the total counts of objects plus the bounding boxes highlighting the detected elements. In this way, they would be able to perform new analysis independently. 
Currently, a first prototype has been designed on a Raspberry Pi 4 \textregistered, but it is still limited in computing capabilities. However, the final goal will be transferring this application on a dedicated server, possibly gpu-supported.


\begin{thebibliography}{}


\bibitem{inproceedings}

Cristoforo, Alessia and Amici, Roberto and Cerri, Matteo and Del Vecchio, Flavia and Hitrec, Timna and Luppi, Marco and Martelli, Davide and Perez, Emanuele and Zamboni, Giovanni, Involvement of serotonergic neurons in thermogenic and wake-promoting effects of orexin-A delivery into the raphe pallidus in the rat, Journal, Volume, page numbers (2016)


\bibitem{hitrec2019neural}
Hitrec, Timna and Luppi, Marco and Bastianini, Stefano and Squarcio, Fabio and Berteotti, Chiara and Martire, Viviana Lo and Martelli, Davide and Occhinegro, Alessandra and Tupone, Domenico and Zoccoli, Giovanna and others, Neural control of fasting-induced torpor in mice, Scientific reports, 9, 1--12 (2019)


\bibitem{lempitsky2010learning}
 Lempitsky, Victor and Zisserman, Andrew,
 Learning to count objects in images, Advances in neural information processing systems,
  1324--1332
  (2010)

\bibitem{segui2015learning}
  Segu{\'\i}, Santi and Pujol, Oriol and Vitria, Jordi,
  Learning to count with deep object features,
  Proceedings of the IEEE Conference on Computer Vision and Pattern Recognition Workshops,
  90--96
  2015

\bibitem{arteta2016counting}
  Arteta, Carlos and Lempitsky, Victor and Zisserman, Andrew,
  Counting in the wild,
  European conference on computer vision,
  483--498
  (2016),

\bibitem{paul2017count}
  Paul Cohen, Joseph and Boucher, Genevieve and Glastonbury, Craig A and Lo, Henry Z and Bengio, Yoshua,
  Count-ception: Counting by fully convolutional redundant counting,
  Proceedings of the IEEE International Conference on Computer Vision,
  18--26
  (2017)



\bibitem{szegedy2016rethinking}
  Szegedy, Christian and Vanhoucke, Vincent and Ioffe, Sergey and Shlens, Jon and Wojna, Zbigniew,
  Rethinking the inception architecture for computer vision,
  Proceedings of the IEEE conference on computer vision and pattern recognition,
  2818--2826,
  2016

\bibitem{cirecsan2013mitosis}
  Cire{\c{s}}an, Dan C and Giusti, Alessandro and Gambardella, Luca M and Schmidhuber, J{\"u}rgen,
  Mitosis detection in breast cancer histology images with deep neural networks,
  International Conference on Medical Image Computing and Computer-assisted Intervention,
  411--418,
  2013,

\bibitem{ciresan2012deep}
  Ciresan, Dan and Giusti, Alessandro and Gambardella, Luca M and Schmidhuber, J{\"u}rgen,
  Deep neural networks segment neuronal membranes in electron microscopy images,
  Advances in neural information processing systems,
  2843--2851,
  2012




\bibitem{AlexNet}
  Alex Krizhevsky and Sutskever, Ilya and Hinton, Geoffrey E,
  ImageNet Classification with Deep Convolutional Neural Networks,
  Advances in Neural Information Processing Systems 25,
  1097--1105,
  2012




\bibitem{YOLO}
    Joseph Redmon and
               Santosh Kumar Divvala and
               Ross B. Girshick and
               Ali Farhadi,
  You Only Look Once: Unified, Real-Time Object Detection,
  CoRR,
  2015



\bibitem{reconstruction}
  Joseph Y. Cheng and
    Feiyu Chen and
    Marcus T. Alley and
    John M. Pauly and
    Shreyas S. Vasanawala,
  Highly Scalable Image Reconstruction using Deep Neural Networks with
               Bandpass Filtering,
  CoRR,



\bibitem{super-resolution}
  Christian Ledig and
    Lucas Theis and
    Ferenc Huszar and
    Jose Caballero and
    Andrew P. Aitken and
    Alykhan Tejani and
    Johannes Totz and
    Zehan Wang and
    Wenzhe Shi,
  Photo-Realistic Single Image Super-Resolution Using a Generative Adversarial
               Network,
  CoRR,
  (2016)



\bibitem{brain_tumor}
  Havaei, Mohammad and Davy, Axel and Warde-Farley, David and Biard, Antoine and Courville, Aaron and Bengio, Yoshua and Pal, Chris and Jodoin, Pierre-Marc and Larochelle, Hugo,
  Brain tumor segmentation with deep neural networks,

  Medical image analysis,
  35,
  18--31
  (2017),


\bibitem{breast_cancer}
  Vandenberghe, Michel E and Scott, Marietta LJ and Scorer, Paul W and S{\"o}derberg, Magnus and Balcerzak, Denis and Barker, Craig,
  Relevance of deep learning to facilitate the diagnosis of HER2 status in breast cancer,

  Scientific reports,
  7,
  45938
  (2017)


\bibitem{torpor_onset},
Oelkrug, Rebecca and Heldmaier, Gerhard and Meyer, Carola,
Torpor patterns, arousal rates, and temporal organization of torpor entry in wildtype and UCP1-ablated mice,
Journal of comparative physiology. B, Biochemical, systemic, and environmental physiology,
181,
137-45
2011





\bibitem{resnet}
  Kaiming He and
               Xiangyu Zhang and
               Shaoqing Ren and
               Jian Sun,
  Deep Residual Learning for Image Recognition,
  CoRR,
  abs/1512.03385,
  (2015)


\bibitem{residual_units}
  Kaiming He and
               Xiangyu Zhang and
               Shaoqing Ren and
               Jian Sun
               ,
  Identity Mappings in Deep Residual Networks,
  CoRR,
  abs/1603.05027,
  (2016)



\bibitem{elastic_tranformation}
P. Y. Simard and D. Steinkraus and J. C. Platt,
Seventh International Conference on Document Analysis and Recognition, 2003. Proceedings,
Best practices for convolutional neural networks applied to visual document analysis,
958-963
(2003)


\bibitem{deep_lab}
L. Chen and G. Papandreou and I. Kokkinos and K. Murphy and A. L. Yuille,
DeepLab: Semantic Image Segmentation with Deep Convolutional Nets, Atrous Convolution, and Fully Connected CRFs
IEEE Transactions on Pattern Analysis and Machine Intelligence,
40,
834--848
(2018)



\bibitem{receptive_field}
 Wenjie Luo and Yujia Li and
 Raquel Urtasun and Richard S. Zemel,
 Understanding the Effective Receptive Field in Deep Convolutional Neural Networks,
 CoRR,
  abs/1701.04128
  (2017),



\bibitem{unet}
  Olaf Ronneberger and
               Philipp Fischer and
               Thomas Brox,
  U-Net: Convolutional Networks for Biomedical Image Segmentation,
  CoRR,
  abs/1505.04597,
  (2015)



\bibitem{gridding}
  Zhengyang Wang and
               Shuiwang Ji,
  Smoothed Dilated Convolutions for Improved Dense Prediction,
  CoRR,
  abs/1808.08931,
  (2018)


\bibitem{deep_resunet}
  Zhengxin Zhang and
               Qingjie Liu and
               Yunhong Wang,
  oad Extraction by Deep Residual U-Net,
  CoRR,
  abs/1711.10684,
  (2017)


\bibitem{inception_code}
  Pavel Yakubovskiy,
  Segmentation Models,
  GitHub repository
  (2019)



\bibitem{Vgg}
Simonyan, Karen and Zisserman, Andrew,
Very Deep Convolutional Networks for Large-Scale Image Recognition,
arXiv 1409.1556
(2014)

\end{thebibliography}

{}










\end{document}